%% file: root.tex
\PassOptionsToPackage{dvipsnames}{xcolor}
\documentclass[letterpaper, 10 pt, conference]{ieeeconf}  

\IEEEoverridecommandlockouts                              

\input{predef}

\title{\LARGE \bf
Large-scale visual SLAM for in-the-wild videos
}

\author{Shuo Sun,$^1$
Torsten Sattler,$^2$
Malcolm Mielle,$^3$
Achim J. Lilienthal,$^{1,4}$
Martin Magnusson$^1$%
\thanks{%
$^1$ AASS research center, Örebro University, Sweden;
$^2$ Czech Technical University in Prague;
$^3$ Independent researcher;
$^4$ Technical University of Munich, Chair: Perception for Intelligent Systems.
This work has received funding from the European Union’s Horizon 2020 research and innovation programme under grant agreement No 101017274 (DARKO) and the Czech Science Foundation under EXPRO
grant UNI-3D (grant no. 23-07973X).
}%
}

\begin{document}

\maketitle
\thispagestyle{empty}
\pagestyle{empty}

\begin{abstract}
  Accurate and robust 3D scene reconstruction from casual, in-the-wild videos can significantly simplify robot deployment to new environments. 
  However, reliable camera pose estimation and scene reconstruction from such unconstrained videos remains an open challenge. Existing visual-only SLAM methods perform well on benchmark datasets but struggle with real-world footage 
which often exhibits uncontrolled motion including rapid rotations and pure forward movements, textureless regions, and dynamic objects.
  We analyze the limitations of current methods and introduce a robust pipeline designed to improve 3D reconstruction from casual videos.
  We build upon recent deep visual odometry methods but increase robustness in several ways.
  Camera intrinsics are automatically recovered from the first few frames using structure-from-motion.
  Dynamic objects and less-constrained areas are masked with a predictive model.
  Additionally, we leverage monocular depth estimates to regularize bundle adjustment, mitigating errors in low-parallax situations.
  Finally, we integrate place recognition and loop closure to reduce long-term drift and refine both intrinsics and pose estimates through global bundle adjustment.  
  We demonstrate large-scale contiguous 3D models from several online videos in various environments. In contrast, baseline methods 
  typically produce locally inconsistent results at several points, producing separate segments or distorted maps.
  In lieu of ground-truth pose data, we evaluate map consistency, execution time and visual accuracy of re-rendered NeRF models.
  Our proposed system establishes a new baseline for visual reconstruction from casual uncontrolled videos found online, demonstrating more consistent reconstructions over longer sequences of in-the-wild videos than previously achieved.
\end{abstract}

\section{Introduction}

Creating 3D maps is a key requirement for most applications of mobile robots, and mature methods exist for accurate mapping with lidar and RGBD data.
However, these methods still require costly hardware and most often skilled staff to calibrate sensors and postprocess results, making the creation of maps and datasets a resource-intensive activity.
If we could create accurate 3D maps from casual ``in-the-wild'' videos found online, that would greatly decrease the deployment effort of mobile robot systems.
Consider for example a tour guide robot for a historical site, where the map could be taken from an existing Youtube video or by unskilled staff casually walking around with a phone, as opposed to surveying the site with a mapping kit.
Reliable 3D scene reconstruction and camera pose estimation from such in-the-wild videos is an open research challenge.
Existing methods for visual-only SLAM (simultaneous localization and mapping) and SfM (structure-from-motion)~\cite{mur2015orb,schonberger2016structure,pan2024global,wang2024vggsfm,duisterhof2024mast3r} work well on benchmark datasets (typically mostly stationary scenes with large camera baselines) but are computationally expensive and struggle in uncontrolled real-world settings, particularly in the presence of large camera rotations, textureless environments, and dynamic objects.
More specifically---as we will show in our experiments and analysis 
(\cref{sec:experiment})---current visual SLAM and SfM methods often fail and generate multiple separate trajectories due to small-parallax motion during the recording and insufficient reliable correspondence estimations across frames.

In this work, we analyze the limitations of current methods 
and introduce a robust pipeline designed to handle these challenges, and push the boundaries of robust 3D scene reconstruction from unconstrained video data.
We primarily focus on robustness,
i.e., we want to generate one consistent trajectory.
In order to work with in-the-wild videos shot with unknown cameras, we first initialize the reconstruction process by estimating camera intrinsics from the first few frames using structure-from-motion.
In \cref{sec:abliation_study} we present an evaluation of several recent methods for recovering focal length and show our method can get accurate focal length for reconstruction.
At the core of our pipeline is the deep visual odometry method DPVO \cite{teed2024deep}, chosen because of its reliable correspondence estimation and efficiency. Instead of keypoint detection and feature matching, DPVO estimates frame-to-frame correspondence by evaluating deep optical flow, which can handle texture-poor regions well.
To handle dynamic objects and unconstrained regions like the sky, we mask them out with a predictive model~\cite{cheng2022masked}.
Unlike traditional methods that rely solely on 2D correspondences---which struggle with large rotations---we use mono-depth estimates \cite{yin2023metric3d} to regularize bundle adjustment (BA), making the pipeline much less sensitive to small-parallax situations, e.g., when the camera is far from the scene 
and rotation-only motion.
Finally, we incorporate place recognition and loop closure to reduce long-term drift. We run a final refinement to refine camera intrinsics, to ensure high-quality results.

Our proposed system establishes a new baseline for visual reconstruction in challenging real-world settings, outperforming existing approaches in scenarios with extreme motion, dynamic elements, and sparse loop closures.
In particular, we demonstrate consistent reconstruction over longer sequences from in-the-wild videos than existing methods can produce.
These reconstruction results may be used for several downstream tasks, such as visual localization,  novel view synthesis, and 3D scene understanding.

In summary, the contributions of this paper are:
\begin{itemize}
    \item 
    Targeting the characteristics of uncalibrated settings, the presence of dynamic objects, small baselines(such as pure rotation), and long distances in the wild videos, we propose a robust visual SLAM system including quick calibration, dynamic object removal, depth-guided BA and pose graph optimization.
    \item 
    Compared to current SOTA SfM methods~\cite{schonberger2016structure,pan2024global}, our method achieves more robust results, generating smooth and continuous trajectories from 15-minute videos.
    \item We propose new metrics to evaluate the robustness and accuracy for in-the-wild video reconstruction, which can work as a baseline for future work.
\end{itemize}

\section{Related Work}
There are many works on 3D reconstruction from 2D RGB images, but the most popular ones share similar pipelines: 1) estimating image correspondence, and 2) optimizing camera poses and scene geometry.
According to the relevance to the paper, we divide the existing work based on whether prior knowledge (specifically, scene geometry estimation) is involved.

\subsection{Without explicit geometry estimation}
Traditional vSLAM~\cite{mur2015orb} and SfM~\cite{schonberger2016structure} methods mainly rely on multiple view tracks to optimize camera poses and geometry. \textit{COLMAP}~\cite{schonberger2016structure} is one prominent example:  
COLMAP begins with image correspondence estimation by detecting and matching SIFT~\cite{lowe2004distinctive} keypoints between images.
After initializing with two-view geometry estimation, COLMAP incrementally registers and triangulates new images until no image is successfully registered.
Visual SLAM can be regarded as an online version with sequential inputs.

Many previous works try to improve the correspondence estimation accuracy by replacing hand-crafted (SIFT) features with learned keypoints~\cite{detone2018superpoint,revaud2019r2d2}.
One can refer to the open repository deep-image-matching\footnote{\href{https://github.com/3DOM-FBK/deep-image-matching}{https://github.com/3DOM-FBK/deep-image-matching}} for more information about modern keypoint detection and matching used in SfM.
These keypoints work well in most cases, but can fail in textureless regions.
Recent detector-free correspondence estimation~\cite{he2024detector} shows better performance on texture-poor regions.
We build our work upon \textit{DPVO}~\cite{teed2024deep} which estimates correspondence via deep optical flow without any keypoints. 

Aside from improving feature detection and matching, another line of work 
focuses on optimization.
Some methods try to improve efficiency by global reconstruction~\cite{pan2024global,theia-manual}, optimizing all frames in one stage.

However, one well-known drawback of current SfM/vSLAM methods is that when the baseline between camera poses is small,  bundle adjustment optimization is unreliable, which often results in scale drift or large errors in pose estimation. 
To reduce the impact of this behavior, it is common to 
filter out points with small triangulation angles~\cite{schonberger2016structure}, which will result in fewer points in the map and make it difficult to register the next image.
In the scenarios considered in this work, where there are (close-to) pure rotations and where the camera can be relatively far from the scene, 
this approach can lead to multiple disjoint reconstructions as the removed points split the trajectory into multiple parts. 

\subsection{With explicit geometry estimation}
Due to the inherent ambiguity in the BA optimization, some prior methods seek to incorporate additional geometry estimates in the BA process.
For example, \textit{StudioSfM}~\cite{liu2022depth} gets the depth image from an extra depth estimator and proposes to add depth regularization terms in the triangulation of TV show scenes with camera movement.
StudioSfm builds upon the incremental SfM COLMAP, but still requires known intrinsics.
The depth is used when registering new images; in our method, we fuse the prior depth in the BA stage to avoid  instability in optimization.
MegaSAM~\cite{li2024megasam}, building on DROID-SLAM~\cite{teed2021droid}, also uses prior depth during the BA optimization.
However, its dense optical flow prediction is not scalable to large-scale scenes.

The recent work Dust3R~\cite{wang2024dust3r} / Mast3R~\cite{leroy2024grounding} and  follow-up works~\cite{elflein2025light3r,duisterhof2024mast3r,murai2024mast3r} can also be regarded as using additional geometry estimation. 
The Dust3R model predicts the 3D point map directly. When aligning multiple frames together, 3D-3D matching is conducted first and followed by 2D-3D refinement.
MASt3R-SLAM~\cite{murai2024mast3r} utilizes Mast3R model prediction directly, tracking new camera frames by aligning point-maps in 3D space.
The dense prediction and multiple frame alignment consume a lot of GPU memory, and cannot be applied to large-scale scenes.
In this work, we focus on long-sequence videos which often consist of thousands of images.
To our knowledge, none of the methods before tried to directly reconstruct from long in-the-wild online videos.


 
\section{Method}\label{sec:method}
As discussed above, current state-of-the-art SLAM/SfM methods often break trajectories into multiple segments when faced with challenging conditions, such as fast rotations commonly found in in-the-wild video sequences.
This paper addresses these limitations by proposing a novel method capable of computing a single cohesive trajectory even under demanding conditions present in casual uncontrolled videos.

Our method is overviewed in \cref{fig:overview}.
We first present our initialization strategy in \cref{sec:method:init}.
In \cref{sec:incremental}, we describe the main reconstruction pipeline,  adopting a deep learning-based optical flow estimation to determine image correspondences within a temporal sliding window, as well as introducing prior depth estimation as an extra regularization term in order to effectively reduce drift errors when the baseline between frames is small.
Furthermore (\cref{sec:loop_closure}), to reduce the drift accumulated during reconstruction, we find loop closures by evaluating  NetVLAD~\cite{arandjelovic2016netvlad} descriptors on the image and optimizing on SIM(3) to fix both scale and pose drift.
Optionally (\cref{sec:post_refinement}), a final 
bundle adjustment step can be performed on the entire sequence to optimize both camera parameters 
and the scene structure.

\input{tables/overview}

\subsection{Initialization}
\label{sec:method:init}

Since we work with in-the-wild videos, camera parameters like focal length are usually unknown. 
Thus, 
we first process a short sequence of frames to set up the scene and to obtain an initial estimate for the 
camera intrinsics.

Assuming a pinhole camera model without distortion, the camera intrinsics can be estimated by selecting $N_{\rm{init}}$ frames that have sufficient difference in optical flow (evaluated by \textit{RAFT}~\cite{teed2020raft} between consecutive frames) in order to get enough parallax for reliable estimation.
Then GLOMAP  is run on the collected images, 
yielding an initial coarse estimate $\mathbf{K}_{\rm{init}}$ of the camera intrinsics (but note that GLOMAP is only used in the initialization stage).
We evaluate the current focal length estimation methods in ~\cref{sec:abliation_study}, our method achieved the most accurate reconstruction though at the cost of some extra seconds.

\subsection{Incremental Reconstruction}\label{sec:incremental}
\subsubsection{Preliminaries}
At the heart of our visual SLAM pipeline, following the acquisition of the initial coarse camera intrinsic parameters $\mathbf{K}_{\rm{init}}$, lies the feature extraction and matching module from \textit{DPVO}~\cite{teed2024deep}.
This module uses a recurrent neural network~(RNN) to estimate correspondences without the need for explicit keypoint detection.

In DPVO, $N_{\rm{patch}}$ patches of $p \times p$ pixels are extracted from each image.
Patch $k$ in frame $i$ is represented by $\mathbf{P}_{ik} = [\mathbf{x}, \mathbf{y}, 1, \mathbf{d}]^{\top}$, where $\mathbf{x},\mathbf{y},\mathbf{d} \in \mathbb{R}^{p^{2} \times 1}$.
Here, $\mathbf{x},\mathbf{y}$ denote the 2D coordinates of the extracted patches, and $\mathbf{d}$ represents the associated inverse depths.
Patches are randomly extracted from each image and each patch is connected to adjacent frames.
When estimating the correspondence of the extracted patch in other frames, feature correlation is conducted between the patch feature and the image feature.
For further details on the neural network architecture used to process the image and estimate correspondence, we point the reader to the original DVPO paper~\cite{teed2024deep}.

Once patches are extracted from the frames and frame-to-frame correspondences have been estimated, bundle adjustment is conducted by minimizing the re-projection error.
The re-projection $\mathbf{P}_{ik}^{j}$ of a given patch $\mathbf{P}_{ik}$ to the frame $j$ is expressed as:
\begin{equation*}
\mathbf{P}_{ik}^{j} = \mathbf{\Pi}_{\rm{init}}(\mathbf{G}_{ij}~ \mathbf{\Pi}_{\rm{init}}^{-1}(\mathbf{P}_{ik}));~ \\
\mathbf{G}_{ij} = \mathbf{G_j} \cdot \mathbf{G}_{i}^{-1}
\end{equation*}
where $\mathbf{\Pi}_{\rm{init}}$ is the camera model constructed from  $\mathbf{K}_{\rm{init}}$, which projects 3D points to 2D pixel coordinates, while $\mathbf{\Pi}_{\rm{init}}^{-1}$ is the inverse projection which re-projects 2D pixels to 3D points in the local coordinate frame.
$\mathbf{G}_{i}$ is the camera pose of frame $i$, representing the \textit{world-to-camera} order.
Assuming that according to the DPVO network prediction, the image correspondence of patch $\mathbf{P}_{ik}$ on frame $j$ is ${\hat{\mathbf{P}}_{ik}^{j}}$, which is the 2D re-projected coordinates and $\hat{\mathbf{P}}_{ik}^{j}=[\hat{\mathbf{x}}^{j},\hat{\mathbf{y}}^{j}] \in \mathbb{R}^{p^{2}\times1}$.
Then the bundle adjustment aims to minimize the re-projection error
\begin{align}
    \mathcal{E}_{(\mathbf{G},\mathbf{d})} = \sum_{(i,j)}\sum_{k}|| \mathbf{P}_{ik}^{j} - \hat{\mathbf{P}}_{ik}^{j} || \label{eqa:ba}
    .
\end{align}

\subsubsection{Semantic Masking}
\label{sec:dynamic_removal}

In casual videos, objects, such as pedestrians or vehicles, frequently appear within the frame.
To avoid potential wrong correspondences caused by dynamic objects, we avoid extracting patches on them. 
Using a semantic segmentation model, areas predicted to contain dynamic objects are masked.
Although semantic masks are a simple strategy, they offer an efficient and direct method for excluding potentially dynamic elements.
To further stabilize 
our optimization, we also prune regions lacking strong constraints, such as \textit{sky}.
 

\subsubsection{Depth-regularized BA}
\label{sec:depth_ba}
Given the frame-to-frame 2D correspondence estimates, we jointly optimize the geometry (i.e., the depth of the patches) and camera poses using bundle adjustment (BA).
As mentioned before, small-parallax views can introduce  ambiguity in depth and pose estimation.
This is why popular SfM implementations~\cite{schonberger2016structure} exclude points lacking sufficient triangulation angles.
Multiple view constraints---i.e. observing the same objects from different positions---can help solve the problem.
However, in uncontrolled walking-tour videos, collecting enough views from varied positions is challenging because the movement (and video stream) 
is mostly going forwards.
On the other hand,  SLAM pipelines using range sensors---such as lidar or RGBD---are not affected by depth ambiguity, even in sequences with pure rotation.

Inspired by recent advances in monocular depth estimation models, we integrate prior depth information into our optimization process.
Specifically, given an image $\mathcal{I} \in \mathbb{R}^{3\times H \times W}$, we input $\mathcal{I}$ and the estimated intrinsic parameters $K_{\rm{init}}$ into the monocular depth estimation model to obtain the corresponding depth map $\mathcal{D} \in \mathbb{R}^{H \times W}$.
When registering the new image $\mathcal{I}_{i}$ into the current reconstruction, we first rescale the depth $\mathcal{D}_{i}$ to align with the existing reconstruction.
The scale factor $\alpha_{i}$ is determined by evaluating $\mathcal{D}_{i}$ with the median depth of the latest three keyframes' patches:
$$
\alpha_i = \frac{\mathrm{median}(\mathcal{D}_{i})}{\mathrm{median}(\mathbf{P}[d]_{{(i-3:i)}})}.
$$
After computing the rescaled depth, we add a depth regularization term to the bundle adjustment to guide the optimization \cref{eqa:ba}:
\begin{align}
    \mathcal{E}_{(\mathbf{G},\mathbf{d})} = \sum_{(i,j)}\sum_{k}|| \mathbf{P}_{ik}^{j} - \hat{\mathbf{P}}_{ik}^{j} || + \mu ||\mathbf{P}_{ik}[d] - \alpha_i \mathcal{D}_{ik} ||, \label{eqa:depth_ba}
\end{align}
where $\mu$ is the regularization term weight.

\subsection{Loop Detection and Pose-graph optimization}
\label{sec:loop_closure}
For large-scale scene reconstruction, accumulation of drift is unavoidable, so place recognition and loop closure are needed to correct the trajectory.
We use NetVLAD~\cite{arandjelovic2016netvlad} to extract feature descriptors $\mathbf{V}_{i}\in\mathbb{R}^{D}$ for each image and, to avoid false positive loop detection, we follow the common practice of requiring 3 consecutive matching frames to register a loop closure; sequences with less than $N$ consecutive frames are disregarded.

When a loop is detected, we run a scale-aware pose graph optimization~\cite{strasdat2010scale}.
This process transform each camera pose from $\mathrm{SE}(3)=\left[\begin{smallmatrix} \mathbf{R} & \mathbf{t} \\ \mathbf{0} &  \mathbf{1}\end{smallmatrix}\right]$ to $\mathrm{SIM}(3)=\left[\begin{smallmatrix} s\mathbf{R} & \mathbf{t} \\ \mathbf{0} &  \mathbf{1}\end{smallmatrix}\right]$ by introducing the scale factor $s \in \mathbb{R}^{+}$.
Given $\Delta S_{ij}$ the transformation between frame $i$ and the detected loop frame $j$, the loop closure residual is defined as:
$$
r = \log_{\mathrm{SIM}(3)}(\Delta S_{ij}^{-1} S_{i} S_{j}),
$$
where $S_{i}$ and $S_{j}$ are the absolute similarity poses.
The pose graph optimization is run synchronously, with the current detected loop frame held fixed while optimizing all previous frames.

\subsection{Post-Refinement}
\label{sec:post_refinement}
Finally, after running the SLAM pipeline on the video sequence,
we perform a post-refinement to get better results.
We first run the feature matching on all images to create a feature database, then we re-triangulate points with achieved camera poses from the SLAM above.
We use the $\texttt{point\_triangulator}$ function in COLMAP to conduct the re-triangulation, enabling refinement of camera intrinsic parameters.
Optionally, we can run $\texttt{ba\_adjuster}$ to refine both camera poses and scene geometry at the cost of time.

\section{Experiments}
\label{sec:experiment}
In this section, we present both quantitative and qualitative experiments demonstrating the robustness of our method on in-the-wild videos.
Our method outperforms current state-of-the-art SfM methods, producing more continuous trajectories with fewer breaks and reduced computation time.
Additionally, we conduct ablation studies to validate the effectiveness of the individual components of our proposed pipeline.

\subsection{Experiment Setup}

\subsubsection{Baselines}
As the baselines, we choose the SOTA structure-from-motion methods COLMAP~\cite{schonberger2016structure} and GLOMAP~\cite{pan2024global}.
We did not compare with some visual SLAM methods~\cite{mur2015orb,teed2021droid} because they are known to be fragile with in-the-wild videos~\cite{patra2019ego}.
For some recent modern SfM methods~\cite{wang2024vggsfm,duisterhof2024mast3r}, due to the heavy GPU requirement, they are not scalable to large-scale scenes.

\subsubsection{Datasets}
We select tour videos (see \cref{tab:overall_res}) from YouTube channels with permissive licenses or with the approval of authors.
We manually remove the ``preview" section from each video, segment the videos into approximately 15-minute clips, and extract frames at 3 FPS to ensure all methods can run within a reasonable timeframe. For drone-recorded videos, we extract frames at 10 FPS due to their higher motion speed.
We rescale images to 512 $\times$ 288, which is compatible with our method. 
For the baselines COLMAP and GLOMAP, we rescale images to 2K resolution because they work better on high-resolution images.

\subsubsection{Evaluation Metrics}
Different from some existing datasets~\cite{sturm2012benchmark,straub2019replica}, we do not have ground-truth camera poses as references for evaluation as we focus on uncontrolled videos.
We propose the following metrics for evaluation: 
\begin{enumerate}
\item The number of  registered images  (counting the images in the largest reconstructed model.)
\item The number of separate models generated (should ideally be one).
\item Breaks along the trajectory. In certain reconstructions, abrupt jumps (indicating significant drift) may occur along the trajectory, without the method recognizing faulty registration and initiating a new model.
Given two consecutive camera positions  $t_i$ and $t_{i+1}$ in the global frame with $\Delta t = ||t_i - t_{i+1}||$, we compute the ratio  $\widehat{\Delta t}_{i} = {||t_i - t_{i+1}||}/{\mathrm{mean}(||\Delta t_{i-k:i+k}||)}$, which means normalizing the position difference based on local scale to remove the scale drift effect.
We define a break if $\widehat{\Delta t}_{i} > 10~\mathrm{mean}({{\Delta t}})$
\item Rendering quality measured by PSNR (peak signal-to-noise ratio). Following the quantitative evaluation done in  \textit{ACE0}~\cite{brachmann2024scene}, 
we build a NeRF model~\cite{tancik2023nerfstudio} along the trajectory and take every eighth view as a test view.
For unregistered images, we set their poses as an identity matrix to penalize incomplete reconstructions.
\end{enumerate}

\subsubsection{Implementation Details}
\label{sec:implementation}
We use \textit{Metric3D}~\cite{yin2023metric3d} for monocular depth estimation and \textit{Mask2Former}~\cite{cheng2022masked} for semantic segmentation.
For the post-refinement, we enable intrinsic refinement when retriangulation.
For  COLMAP/GLOMAP, we apply SIFT feature matching, each image is matched with 20 frames before and after, and we enable vocabulary tree matching to allow loop detection.
COLMAP requires a lot of time when running on thousands of images. In our experiments, we adopt the fast version parameters\footnote{\href{https://github.com/colmap/colmap/issues/116}{https://github.com/colmap/colmap/issues/116}} when running COLMAP.

\subsection{Quantitative Results}
We demonstrate the robustness of our method in \cref{tab:overall_res}. 
%
%
In all video sequences, our method registers the most images in each sequence without breaking the scene into multiple models. 
In contrast, COLMAP can generate multiple models due to failed image registration.
Though GLOMAP also always generates one single model, there are often multiple breaks in the generated results, i.e., failed registrations that are not detected by the method.
Note that our method misses two images in the sequence of ``Uppsala". 
This is due to our post-refinement stage, 
where two images fail to generate enough SIFT feature correspondences. 
%
It is also worth mentioning that, though we use a fast configuration of COLMAP, its running time is much longer than ours.
Our method is both more robust and more efficient. 

If the 3D map has been accurately reconstructed from the input video, it should be possible to create a NeRF model from the registered frames and compute the rendering quality of a held-out image to the corresponding NeRF rendering.
Since plain NeRF does not have enough capacity for outdoor unbounded large-scale scenes, we cut the long sequences into short sequences with $500$ frames each.
Then we build a NeRF model~(specifically, \href{https://docs.nerf.studio/nerfology/methods/nerfacto.html}{Nerfacto}) on each short sequence, with every 8th frame held out and used as a test frame. 
The novel view synthesis performance for the test frames 
is reported in \cref{tab:psnr}.
Our method achieves better rendering results due to fewer breaks in the trajectory.
When COLMAP or GLOMAP aligns well, their estimated camera poses are more accurate.  However, our method might be less precise, but more robust overall.
As shown in \cref{tab:he-psnr-clipped}, for the sequence ``Helsingborg Seq-1", COLMAP and GLOMAP register images well for the first 500 frames,  and achieve better rendering results; while in frame 500--1000, COLMAP and GLOMAP have breaks in the trajectory, resulting in low rendering results.
\input{tables/he_1_two_sequences}
As shown in \cref{fig:short_seq_details}, there is a break in the trajectory generated by COLMAP.
In the NeRF evaluation (\cref{fig:break_rendering}), we can see the resulting rendering artifacts at the break pose.
\input{tables/short_seq_compare}%
\input{tables/break_rendering}It should be noted that all reported PNSR values are rather low, in part because the trajectories generated by the methods are not perfect but also because the NeRF model used to evaluate struggles with the large-scale outdoor scenes~\cite{tancik2022block}.

\input{tables/model_eval}

\input{tables/rendering_result}

\subsection{Qualitative Results}
In addition to the quantitative evaluations above, we also compare the recreated paths with approximate GPS tracks from the data sets, where available. 
As shown in \cref{fig:qualtative_demonstration}, our method can achieve smooth and consistent trajectories.
Though GLOMAP can register more frames than COLMAP (as seen in \cref{{tab:overall_res}}), the resulting path and 3D model is often inconsistent for these kinds of uncontrolled input videos.
We do not show ``Taicang Park" and ``The Backyard" sequence because all methods perform well on these two ones.

\input{tables/qualtative_res}

\subsection{Ablation Studies}
\label{sec:abliation_study}
We conducted experiments to demonstrate the effectiveness of the different parts of our proposed reconstruction pipeline.
\cref{fig:abliation_study} shows the reconstruction results for the ``Yanshan Park" sequence with and without depth regularization, masking dynamic objects, and loop closure.
Referring to the reconstruction of COLMAP and GLOMAP on the same sequence in \cref{fig:qualtative_demonstration}, as we said, when COLMAP successfully registers images, the overall result is reliable.
Based on this comparison and checking the video, we can say our method achieves good and accurate reconstruction on this sequence.
\input{tables/abliation_study}

\begin{figure}[t]
    \centering
    \begin{subfigure}{0.23\textwidth}
        \includegraphics[width=\linewidth]{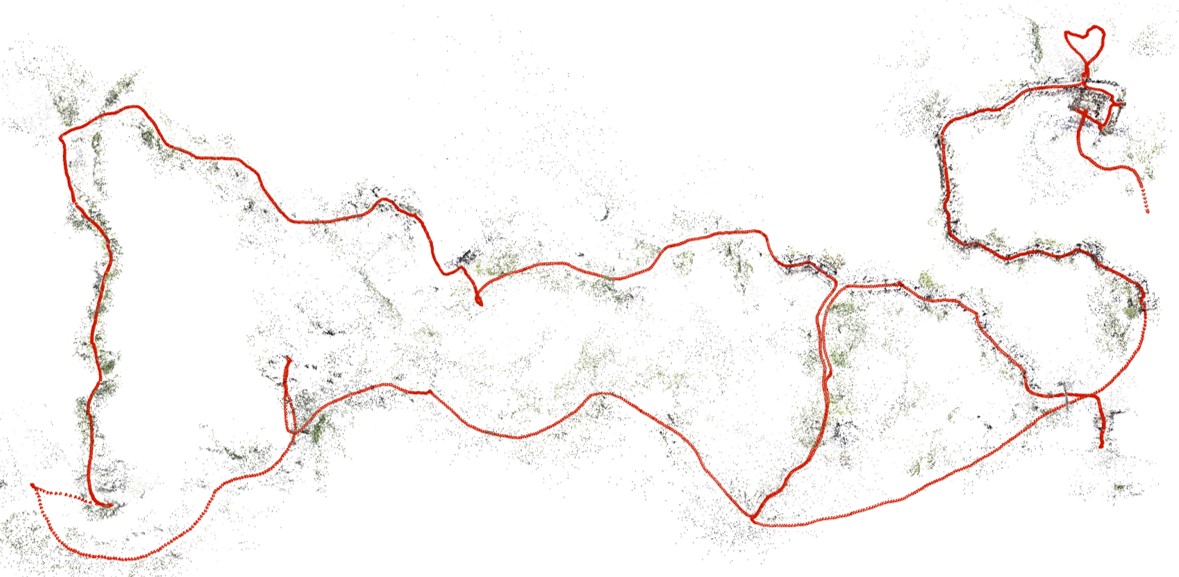}
        \caption{MoGE~($f=396.69$)}
    \end{subfigure}
    \hfill
    \begin{subfigure}{0.23\textwidth}
        \includegraphics[width=\linewidth]{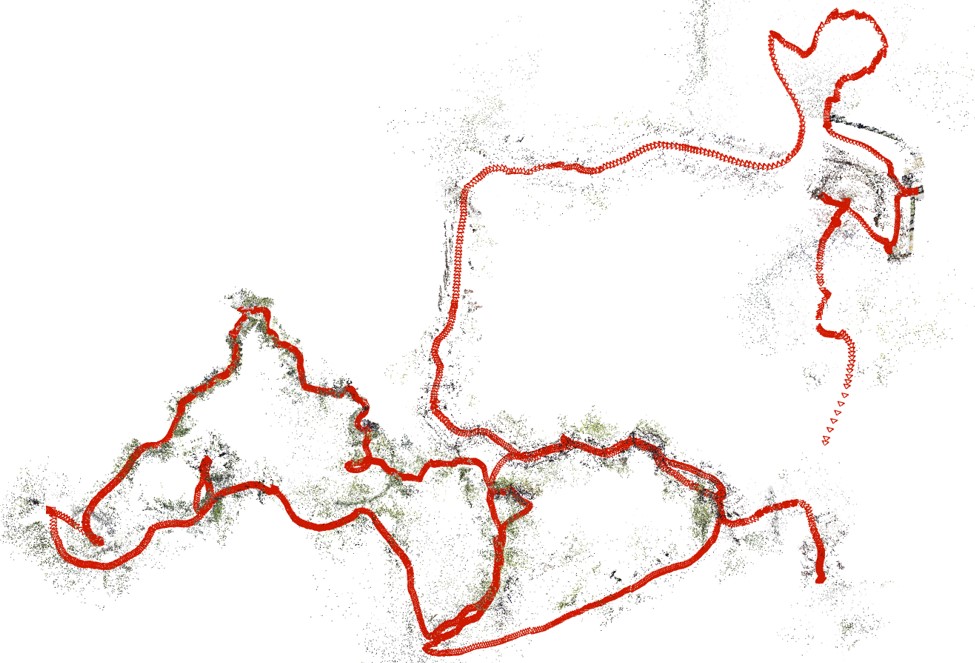}
        \caption{Mast3R~($f=338.61$)}
    \end{subfigure}
    \vspace{0.1cm}

    \begin{subfigure}{0.23\textwidth}
        \centering\includegraphics[width=0.85\linewidth]{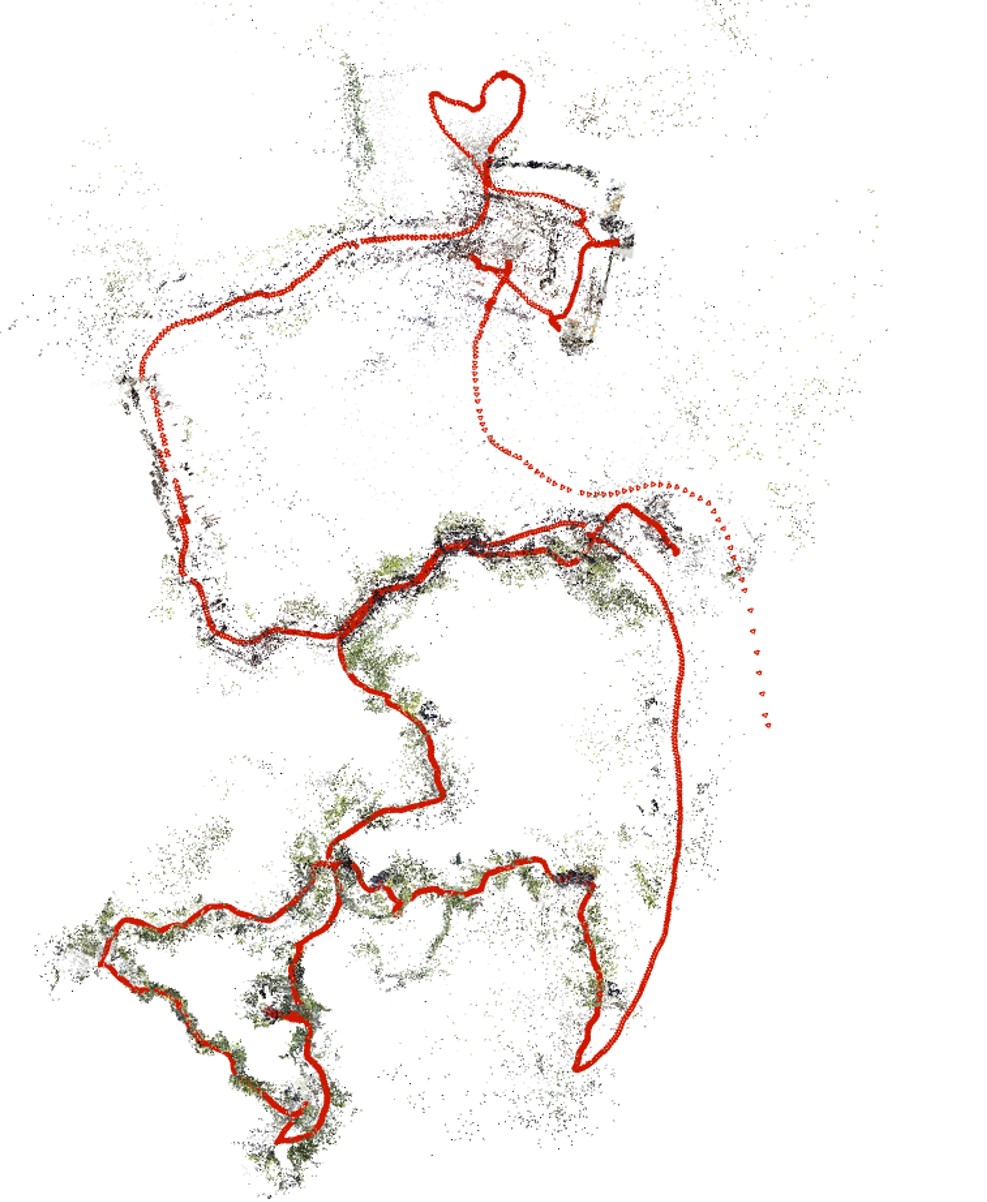}\\
        \caption{GeoCalib~($f=374.3\pm102$)}
    \end{subfigure}
    \hfill
    \begin{subfigure}{0.23\textwidth}
        \includegraphics[width=\linewidth]{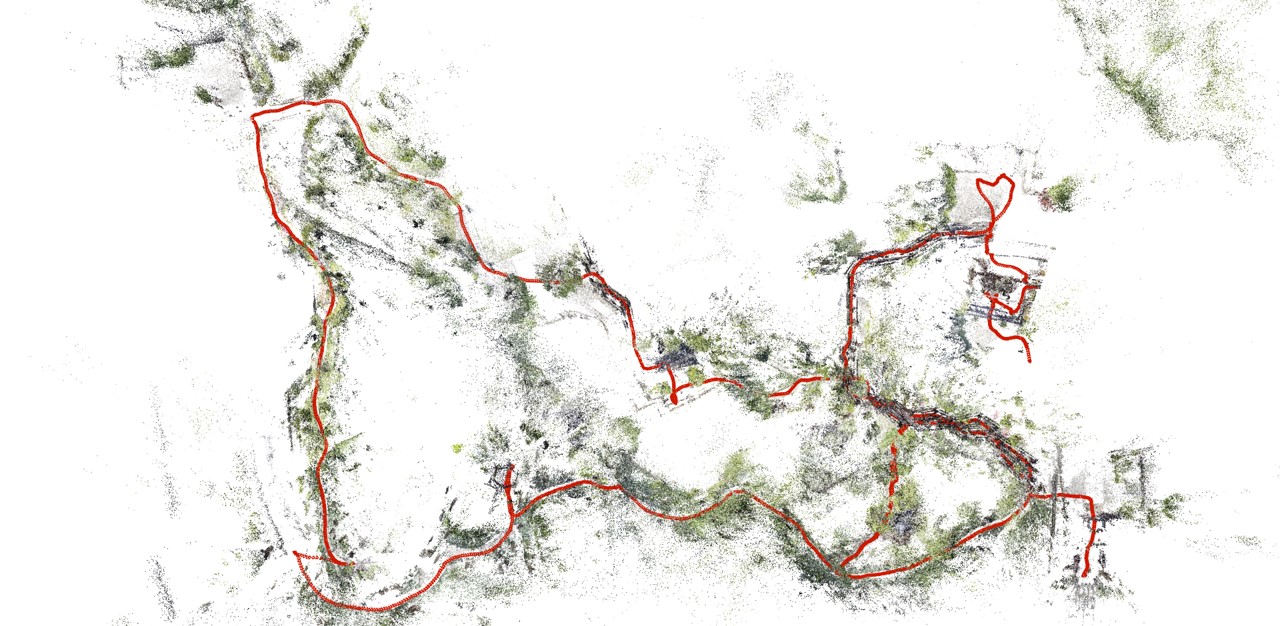}
        \caption{Ours~($f=409.89$)}
    \end{subfigure}
    \caption{Reconstruction results by different ways of estimating focals.}
    \label{fig:focal_diff}
\end{figure}

We also tested current camera intrinsic parameter estimation methods.
We tried \textit{Mast3R}~\cite{leroy2024grounding}, \textit{MoGE}~\cite{wang2024moge} and \textit{GeoCalib}~\cite{veicht2024geocalib}, where Mast3R and MoGE recover focal length from the predicted point cloud while GeoCalib directly predicts from the image.
We select the first image (first two for Mast3R) in the video and feed it into the above-mentioned methods.
We assume a pinhole camera model where the principal point is in the center.
For the sequence ``{Yanshan Park}", the estimated focal lengths and corresponding reconstruction results are presented in \cref{fig:focal_diff}.
Locally, imprecise intrinsic parameter estimation will not result in a severe failure, but with longer trajectories, it will incur more drifts.
Compared to the aforementioned methods that rely on only one or two images to estimate the focal length, our method is more computationally expensive but achieves higher accuracy, making it particularly beneficial for long sequences.

\section{Conclusions}
Robust and accurate 3D reconstruction from in-the-wild videos is a very challenging problem.
Compared to standard SLAM datasets, where the camera is typically carefully moved through a scene to ensure being able to track its movement, there is no control over the camera motion and we often observe pure rotations or pure forward motion, which are challenging. At the same time, there often are moving objects in the scene, complicating the process. 
Addressing these challenges represents the next frontier in SLAM, and progress in this direction will lead to more robust and adaptable systems, crucial for real-world robotics applications.

With the present work we take strides towards consistent 3D mapping of large-scale environments from uncontrolled videos: over 1~km in length, thousands of video frames, over 10~min duration.
Specifically, we have investigated robust methods for recovering the focal length from in-the-wild videos, leverage semantic masks to improve data association in scenes with moving objects, and use monocular depth cues to regularize bundle adjustment in order to be more robust to difficult camera motion and features near the horizon. Comparing our pipeline to GLOMAP~\cite{pan2024global} and COLMAP~\cite{schonberger2016structure}, our proposed method robustly produces longer sequences without breaks, and does so in a fraction of the time.

Future research directions include methods for further reducing drift while maintaining consistent reconstructions and strategies for cases where the view is severely covered by moving objects.




\printbibliography
\end{document}

%% file: predef.tex
\usepackage[font=small,labelfont=bf]{caption}
\usepackage{booktabs}
\usepackage{multirow}
\usepackage{colortbl}
\usepackage{array}
\usepackage{todonotes}

\usepackage{pifont}
\usepackage{amssymb}
\usepackage{amsmath}

\usepackage{csquotes}
\usepackage[breaklinks,colorlinks=true,linkcolor=red, citecolor=blue]{hyperref}%
\usepackage[capitalise]{cleveref}
\usepackage{longtable}
\usepackage[hyperref=true, backref=false, sorting=none,backend=biber]{biblatex}

\usepackage{arydshln}
\usepackage{graphicx}
\usepackage{subcaption}
\usepackage{adjustbox}
\usepackage{bbding}
\usepackage{makecell}

\colorlet{colorFst}{Green!40}       
\colorlet{colorSnd}{SpringGreen!50} 
\colorlet{colorTrd}{Yellow!40}      
\colorlet{colorLow}{darkgray!30}    
\newcommand{\fs}{\cellcolor{colorFst}\bf}   
\newcommand{\nd}{\cellcolor{colorSnd}}      
\newcommand{\rd}{\cellcolor{colorTrd}}      

\makeatletter
\def\footnoterule{\kern-3\p@
  \hrule \@width 2in \kern 2.6\p@} 
\makeatother

\AtBeginBibliography{\scriptsize}

%% file: tables/overview.tex
\begin{figure*}[ht]
    \centering
    \includegraphics[width=0.85\linewidth]{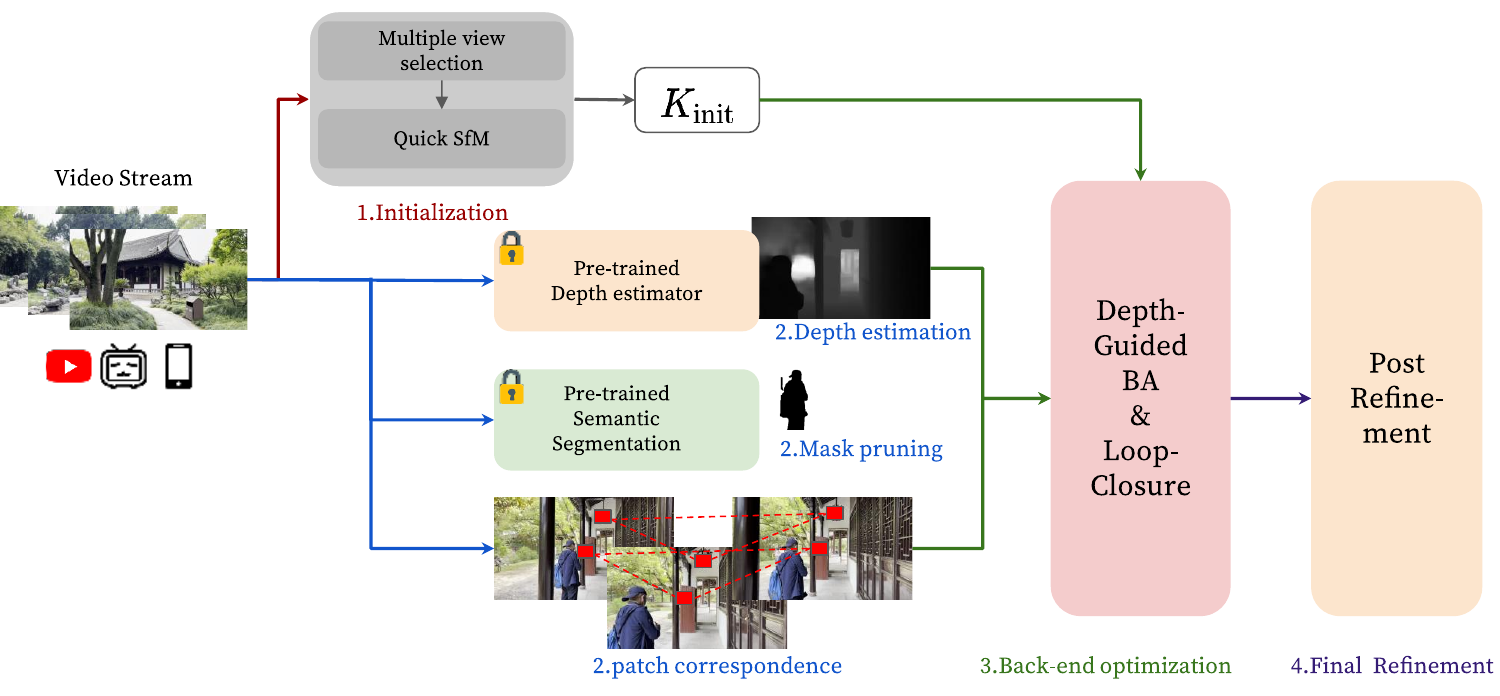}
    \caption{\textbf{Overview of our method}.
    Given a video stream, we extract frames from the video sequentially.
    We first run an efficient global SfM process to estimate the camera intrinsic parameters $K_{\mathrm{init}}$ (\cref{sec:method:init}).
    Using an off-the-shelf semantic segmentation model, we prune the potential objects in the image (\cref{sec:dynamic_removal}) when estimating correspondence between frames.
    Correspondences are estimated across frames by DPVO; we use a monocular depth estimation model to get the prior depth, which can be used to guide the bundle adjustment optimization to improve robustness (\cref{sec:depth_ba}).
    SIM(3) pose graph optimization is conducted if a loop closure is detected (\cref{sec:loop_closure}).
    Finally, we run a re-triangulation and optionally global BA to refine the camera parameters and scene geometry (\cref{sec:post_refinement}).}
    \label{fig:overview}
\end{figure*}

%% file: tables/he_1_two_sequences.tex
\begin{table}
\centering
\caption{Rendering results on selected sequences. The reported results are the average/min/max PSNR values computed for segments of 500 frames.
\label{tab:psnr}
}
\begin{tabular}[t]{lcccc}

\toprule
\textbf{Sequence} & \textbf{Metrics} & \textit{COLMAP}~\cite{schonberger2016structure} & \textit{GLOMAP}~\cite{pan2024global} & \textit{Ours} \\

\midrule

\multirow{4}{*}{Yanshan Park} 
& Avg & \nd 13.64 & \rd 13.35 & \fs 13.95\\
&  min & \nd 12.12 & \rd 10.32 & \fs 12.51 \\
&  max &  \nd 14.48 & \rd 15.08 & \fs 15.55 \\

\hdashline

\multirow{4}{*}{Taicang Park} 
& Avg & \rd 16.68 & \nd 16.77 & \fs 17.14 \\
&  min & \nd 13.36 & \rd 12.79 & \fs 15.64 \\
&  max &  \rd 18.67 & \fs 18.97 & \nd 18.73 \\

\hdashline

\multirow{4}{*}{Uppsala} 
& Avg & \rd 12.71 & \nd 14.77 & \fs 15.25 \\
&  min & \rd 7.7 & \nd 12.13 & \fs 14.43 \\
&  max &  \rd 14.83  & \fs 19.75 & \nd 17.62 \\

\hdashline

\multirow{4}{*}{Helsingborg-1} 
& Avg & \fs 15.18 & \rd 12.61 & \nd 14.72 \\
&  min & \nd 13.01 & \rd 9.26 & \fs 13.66 \\
&  max & \fs 16.47 & \rd 16.22 & \nd 15.77 \\

\hdashline

\multirow{4}{*}{Helsingborg-2} 
& Avg & \rd 14.21 & \nd 14.56 & \fs 14.60 \\
&  min & \rd 10.86 & \nd 12.18 & \fs 13.17 \\
&  max & \fs 17.74 & \rd 17.18 & \nd 17.68 \\

\hdashline

\multirow{4}{*}{Lund} 
& Avg & \rd 12.74 & \nd 13.04 & \fs 13.75 \\
&  min & \rd 11.7 & \nd 12.06 & \fs 13.16 \\
&  max &  \rd 13.99 & \nd 14.22 & \fs 14.28 \\

\bottomrule

\end{tabular}
\end{table}

%% file: tables/short_seq_compare.tex
\begin{figure}
    \centering
    \begin{subfigure}{0.08\textwidth}  
        \centering
        \includegraphics[width=\textwidth]{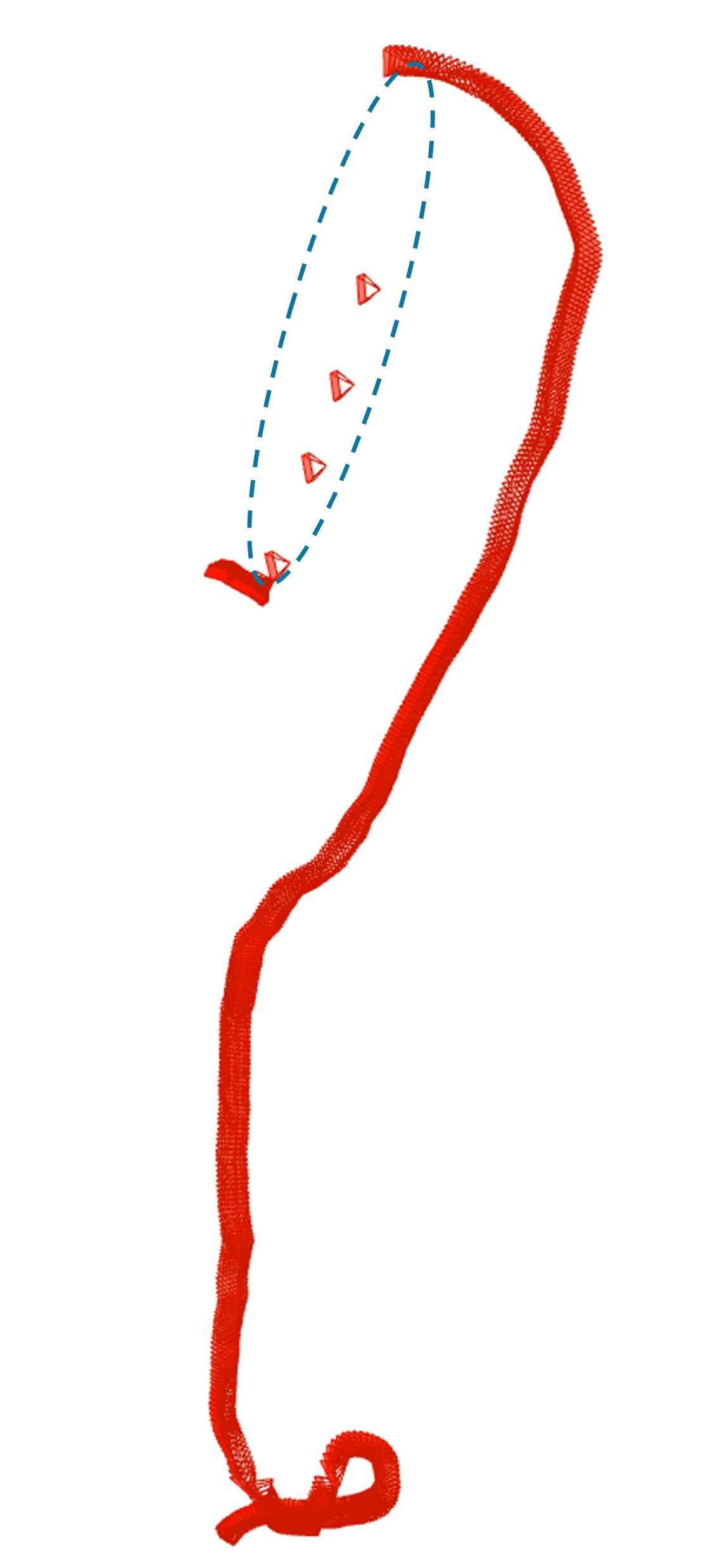}
        \caption{\fontsize{7bp}{12bp}\textit{COLMAP}}
        \label{fig:subfig1}
    \end{subfigure}
    \hfill
    \begin{subfigure}{0.08\textwidth}  
        \centering
        \includegraphics[width=\textwidth]{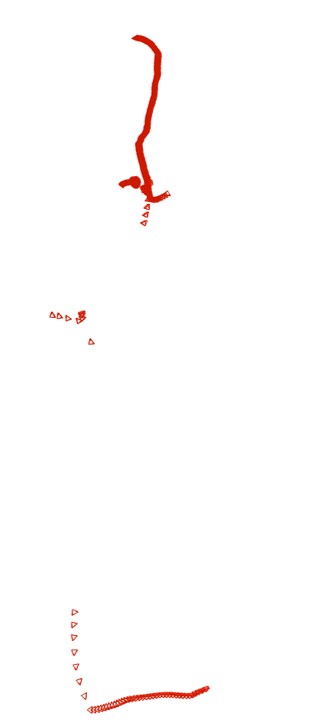}
        \caption{\fontsize{7bp}{12bp}\textit{GLOMAP}}
        \label{fig:subfig1}
    \end{subfigure}
    \hfill
    \begin{subfigure}{0.08\textwidth}  
        \centering
        \includegraphics[width=\textwidth]{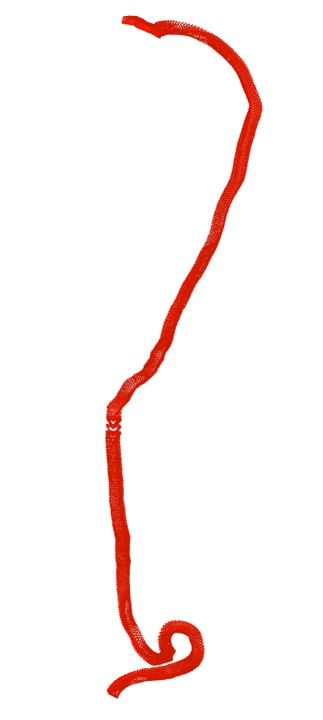}
        \caption{\fontsize{7bp}{12bp}\textit{Ours}}
        \label{fig:subfig1}
    \end{subfigure}

    \caption{The camera poses ({\color{red}{red}} in the map) on the sequence ``Helsingborg  Seq-1" across frames 500--1000.
    COLMAP has a break in the trajectory (circled in {\color{blue}{blue}}); GLOMAP tacitly fails to register images; our method produces smooth and continuous trajectories.}
    \label{fig:short_seq_details}
\end{figure}

%% file: tables/break_rendering.tex
\begin{figure}
    \centering
    \begin{subfigure}{0.23\textwidth}  
        \centering
        \includegraphics[width=\textwidth]{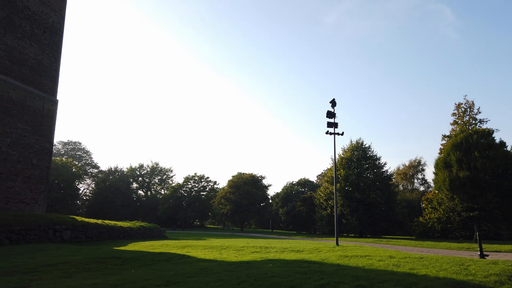}
        \caption{Reference Image}
        \label{fig:subfig1}
    \end{subfigure}
    \hfill
    \begin{subfigure}{0.23\textwidth}
        \centering
        \includegraphics[width=\textwidth]{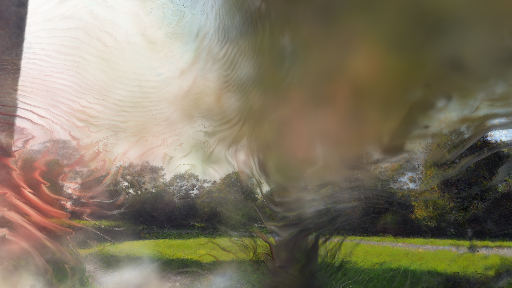}
        \caption{{COLMAP}~(PSNR:8.25)}
        \label{fig:subfig2}
    \end{subfigure}
    
    \vspace{0.2cm}

    \begin{subfigure}{0.23\textwidth}
        \centering
        \includegraphics[width=\textwidth]{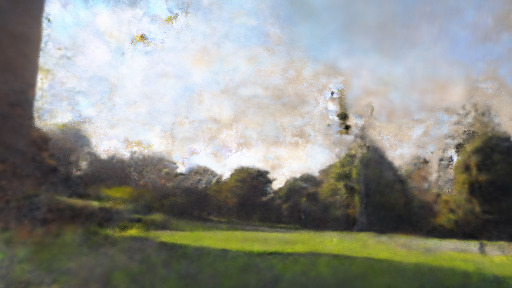}
        \caption{{GLOMAP}~(PSNR:14.16)}
        \label{fig:subfig3}
    \end{subfigure}
    \hfill
    \begin{subfigure}{0.23\textwidth}
        \centering
        \includegraphics[width=\textwidth]{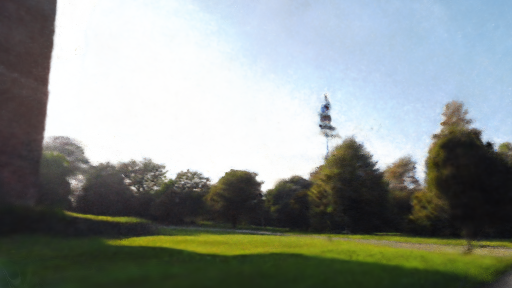}
        \caption{{Ours}~(PNSR:16.80)}
        \label{fig:subfig4}
    \end{subfigure}

    \caption{The rendering results at the camera pose where COLMAP breaks.}
    \label{fig:break_rendering}
\end{figure}

%% file: tables/model_eval.tex
\begin{table*}
\urlstyle{same}
\centering
\caption{Reconstruction results on in-the-wild videos.
Our method achieves most robust performance while using the least time for long sequence videos.
}
\begin{tabular}[t]{p{0.3\linewidth}clllccc}

\toprule
\textbf{Sequence} & \textbf{Screenshot} & & \textbf{\#frames} & \textbf{Metrics} & \textit{COLMAP}~\cite{schonberger2016structure} & \textit{GLOMAP}~\cite{pan2024global} & \textit{Ours} \\

\midrule

\multirow{4}{*}{\parbox{\linewidth}{
Yanshan Park, China~
  \\\url{https://youtu.be/D8B30GIX-8s}
  }}
& \multirow{4}{*}{\adjustbox{valign=c,width=0.11\linewidth}{\includegraphics{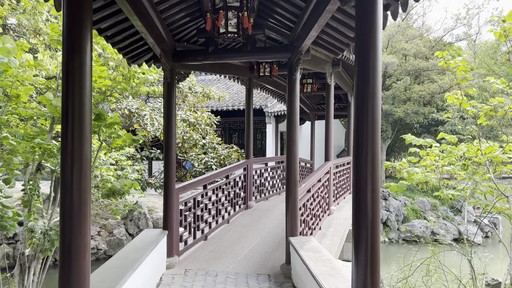}}} & & \multirow{4}{*}{3327}& \#~Registered & \nd2989 & \fs3327 & \fs3327 \\
& & & & \#~Models & \nd2 & \fs1 & \fs1 \\
& & & & \#~Breaks &  \nd3 &  \rd5 & \fs0 \\ 
& & & & Time(min) & \rd665 & \nd149 & \fs18 \\

\hdashline
\multirow{4}{*}{\parbox{\linewidth}{
  Taicang Park,
  China
  \\\url{https://youtu.be/LJf7LKLvmUc}}}
  & \multirow{4}{*}{\adjustbox{valign=c,width=0.11\linewidth}{\includegraphics{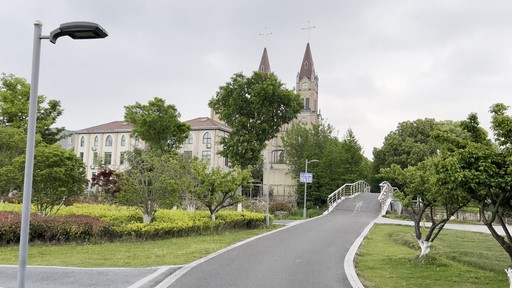}}} & & \multirow{4}{*}{2597}& \#~Registered &  \rd 2534 & \nd 2597 & \fs 2597 \\
& & & & \#~Models & \nd 2 & \fs 1 & \fs 1\\
& & & &  \#~Breaks & \rd 10 & \nd 1 & \fs 0\\
& & & &  Time(min) & \rd 385 & \nd 183 & \fs 8 \\

\hdashline
\multirow{4}{*}{\parbox{\linewidth}{
  Uppsala, Sweden,
  \\\url{https://youtu.be/aVh_jTIP2cE?t=1262}}} & \multirow{4}{*}{\adjustbox{valign=c,width=0.11\linewidth}{\includegraphics{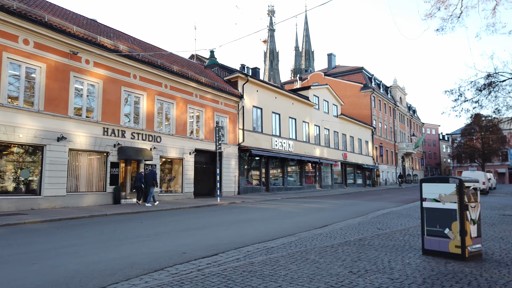}}} & & \multirow{4}{*}{2533}& \#~Registered &  \rd2206 & \nd 2528 & \fs 2531 \\
& & & & \#~Models & \nd3 & \fs1 & \fs1\\
& & & & \#~Breaks & \nd2 & \rd3 & \fs1 \\
& & & & Time(min) & \rd200 & \nd120 & \fs12 \\

\hdashline


\multirow{4}{*}{\parbox{\linewidth}{
  Nanxun Ancient Town, China
  \\\url{https://youtu.be/Owukwe_8OGw}}}& \multirow{4}{*}{\adjustbox{valign=c,width=0.11\linewidth}{\includegraphics{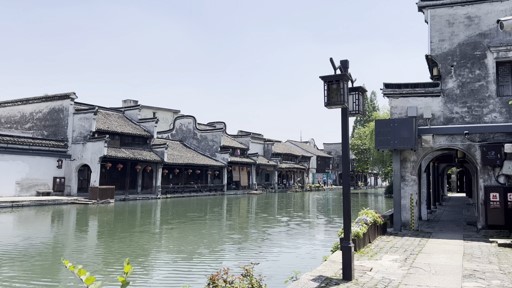}}} & & \multirow{4}{*}{1026} & \#~Registered & \fs1026 & \fs1026 & \fs1026 \\
& & & & \#~Models & \fs1 & \fs1 & \fs1 \\
& & & & \#~Breaks & \fs0 & \fs0 & \fs0 \\
& & & & Time(min) & \rd57 & \rd30 & \fs6 \\

\hdashline
\multirow{8}{*}{\parbox{\linewidth}{
  Helsingborg, Sweden
  \\\url{https://youtu.be/wUZ_zslH3vY?t=300} 
  \\\url{https://youtu.be/wUZ_zslH3vY?t=1200}}}& \multirow{8}{*}{\adjustbox{valign=c,width=0.11\linewidth}{\includegraphics{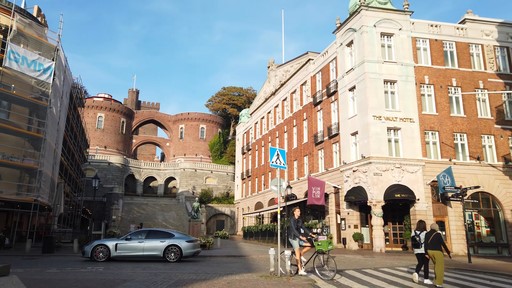}}} & \multirow{4}{*}{Seq-1} & \multirow{4}{*}{2700}& \#~Registered & \rd2381 & \nd2382 & \fs2700 \\
& & & & \#~Models & \nd2 & \fs1 & \fs1 \\
& & & & \#~Breaks & \nd3 & \rd29 & \fs0 \\
& & & & Time(min) & \rd303 & \nd154 & \fs16 \\
\cdashline{3-8}
& & \multirow{4}{*}{Seq-2} & \multirow{4}{*}{2700} & \#~Registered & \nd2689 & \rd2279 & \fs2700\\
& & & & \#~Models & \nd2 & \fs1 & \fs1 \\
& & & & \#~Breaks & \nd1 & \rd13 & \fs0 \\
& & & & Time(min) & \rd258 & \nd140 & \fs18 \\

\hdashline
\multirow{4}{*}{\parbox{\linewidth}{
  Lund, Sweden
  \\\url{https://youtu.be/Nhc5BNlfDms?t=1800}}} & \multirow{4}{*}{\adjustbox{valign=c,width=0.11\linewidth}{\includegraphics{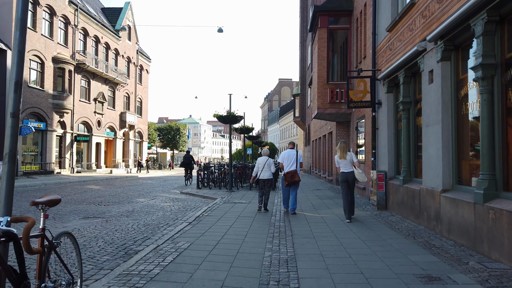}}} & & \multirow{4}{*}{2700}& \#~Registered & \rd1437 & \nd2697 & \fs2700 \\
& & & & \#~Models & \nd3 & \fs1 & \fs1 \\
& & & & \#~Breaks & \nd1 & \rd17 & \fs0 \\
& & & & Time(min) & \rd300 & \nd180 & \fs16 \\

\hdashline
\multirow{4}{*}{
\parbox{\linewidth}{
  The Backyard, USA
  \\\url{https://youtu.be/OtkZJbW_sO0}}}
  & \multirow{4}{*}{\adjustbox{valign=c,width=0.11\linewidth}{\includegraphics{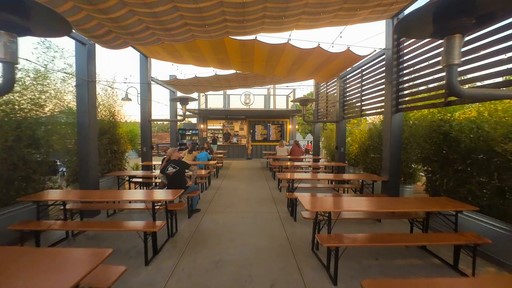}}} &  & \multirow{4}{*}{578}&  \#~Registered & \nd577 & \nd577 & \fs578 \\
& & & & \#~Models & \fs1 & \fs1 & \fs1\\
& & & & \#~Breaks & \fs0 & \nd1 & \fs0 \\
& & & & Time(min) & \nd15 & \fs4 & \fs4 \\

\hdashline


\textbf{Time Average} (min) & & & & & \rd2169 & \nd956 & \fs12\\
\bottomrule

\end{tabular}
\label{tab:overall_res}
\end{table*}%

%% file: tables/rendering_result.tex
\begin{table}
\centering
\caption{NeRF rendering results on smaller clipped parts on ``Helsingborg  Seq-1".
\label{tab:he-psnr-clipped}
}
\begin{tabular}[t]{lccc}

\toprule
\textbf{Sequence} & \textit{COLMAP}~\cite{schonberger2016structure} & \textit{GLOMAP}~\cite{pan2024global} & \textit{Ours} \\

\midrule

frame 0--500 & \fs16.41 & \nd16.22 & \rd14.05\\
frame 500--1000 & \nd14.99 & \rd9.26 & \fs15.74\\

\bottomrule

\end{tabular}
\end{table}

%% file: tables/qualtative_res.tex
\begin{figure*}
\centering
{
\footnotesize
\setlength{\tabcolsep}{1pt}
\renewcommand{\arraystretch}{1}
\newcommand{\sz}{0.32}
\newcommand{\figsize}{4.2cm}
\newcommand{\smallfigsize}{3.9cm}
\begin{tabular}{cm{\figsize}m{\figsize}m{\figsize}m{\smallfigsize}}

\toprule

\textbf{Sequence} & ~~~~~~~~~~~~~~~~\textit{COLMAP}~\cite{schonberger2016structure} & ~~~~~~~~~~~~~~~~\textit{GLOMAP}~\cite{pan2024global} & ~~~~~~~~~~~~~~~~\textit{Ours} & ~~~~~~~~~~~~~~~~Reference\\

\midrule


\rotatebox[origin=c]{90}{Yanshan Park} & \includegraphics[width=\linewidth]{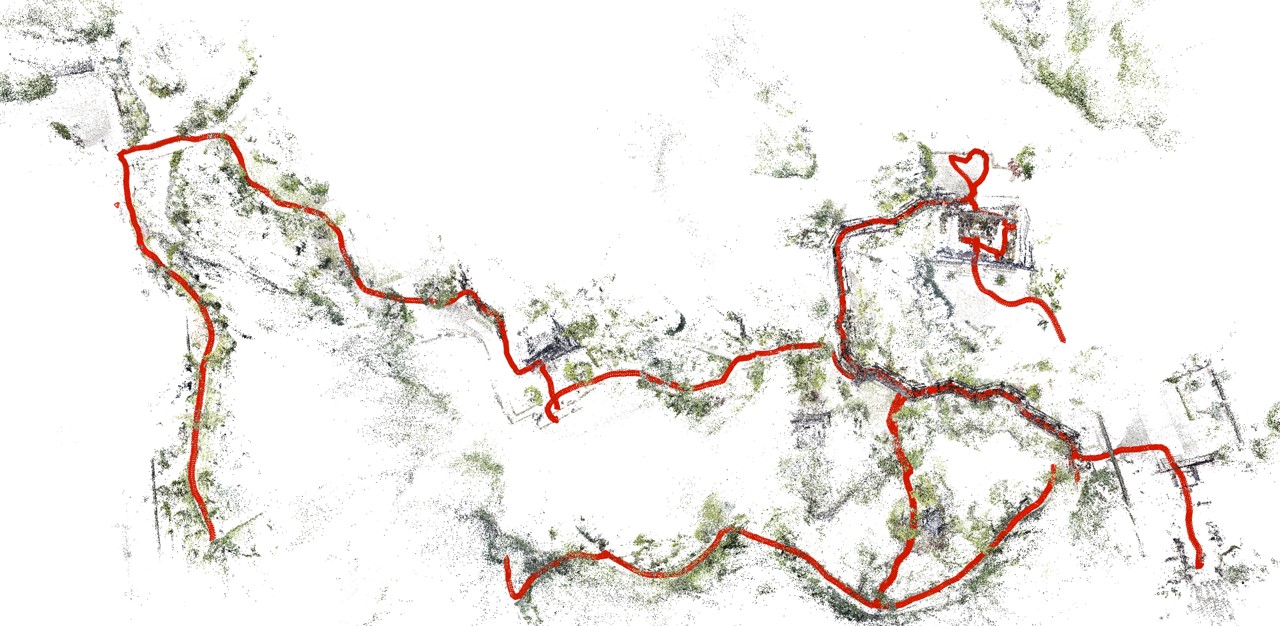}& \includegraphics[width=\linewidth]{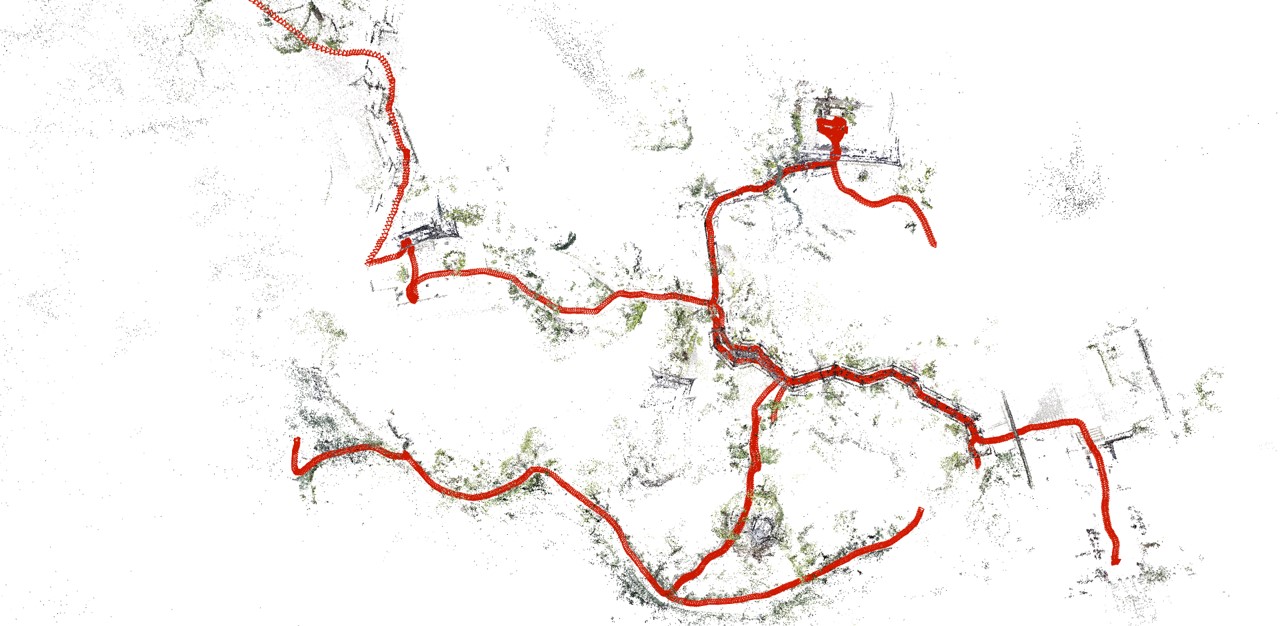}&
\includegraphics[width=\linewidth]{figures/yanshanyuan/ours_park2.jpg} &
~~~~~~~~~~~~~~~~None
\\

& ~~~~~~~~~~~~~~~~disjoint & ~~~~~~~~~~~~frames missed~\&~disjoint & & \\

\hdashline

\rotatebox[origin=c]{90}{Helsingborg-1} & \includegraphics[width=\linewidth]{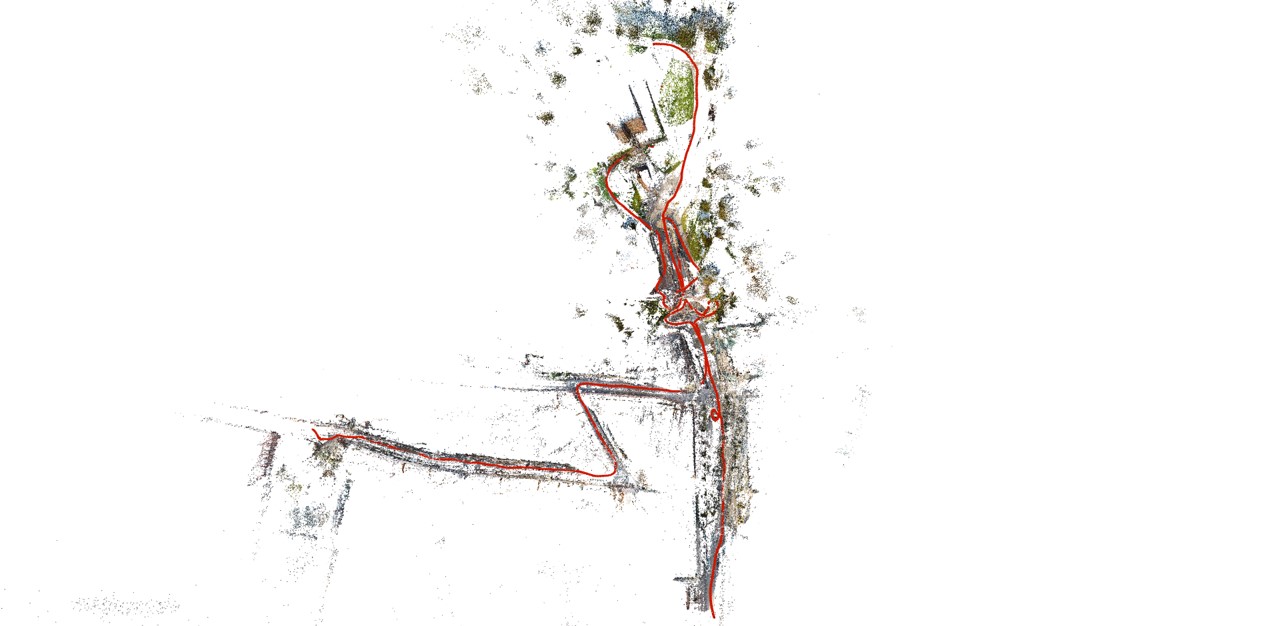}& \includegraphics[width=\linewidth]{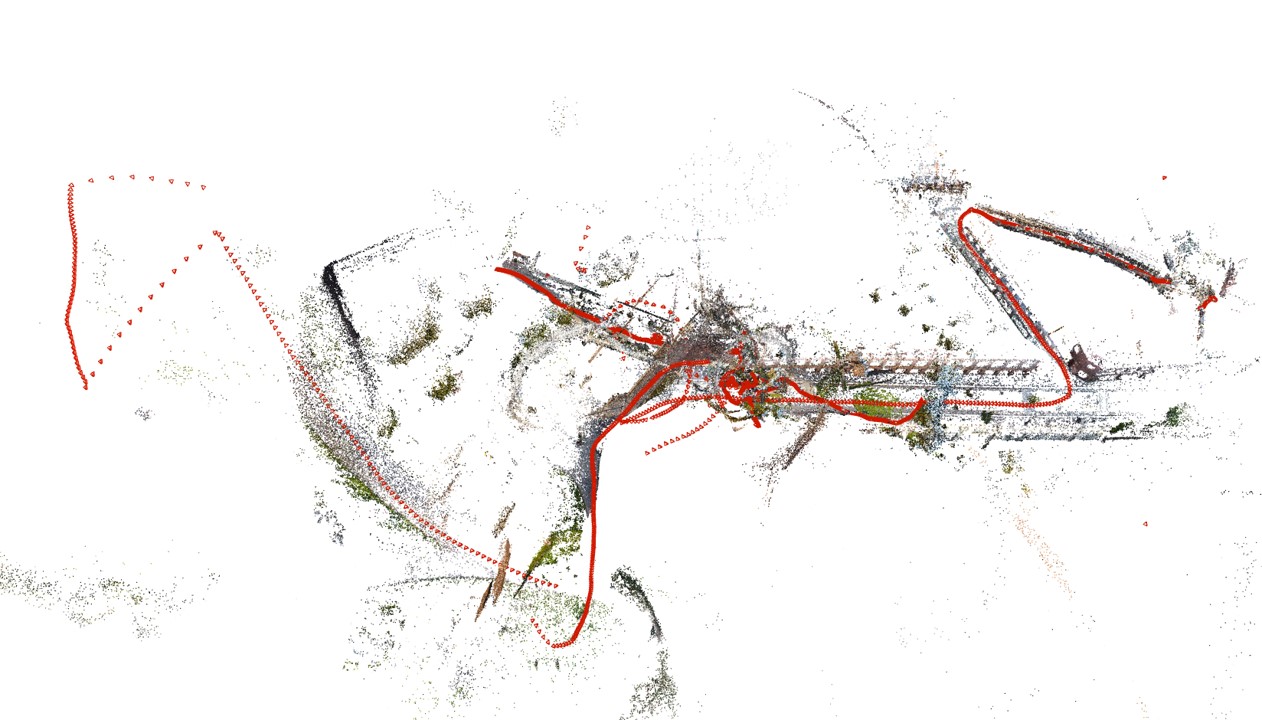}&
\includegraphics[width=\linewidth]{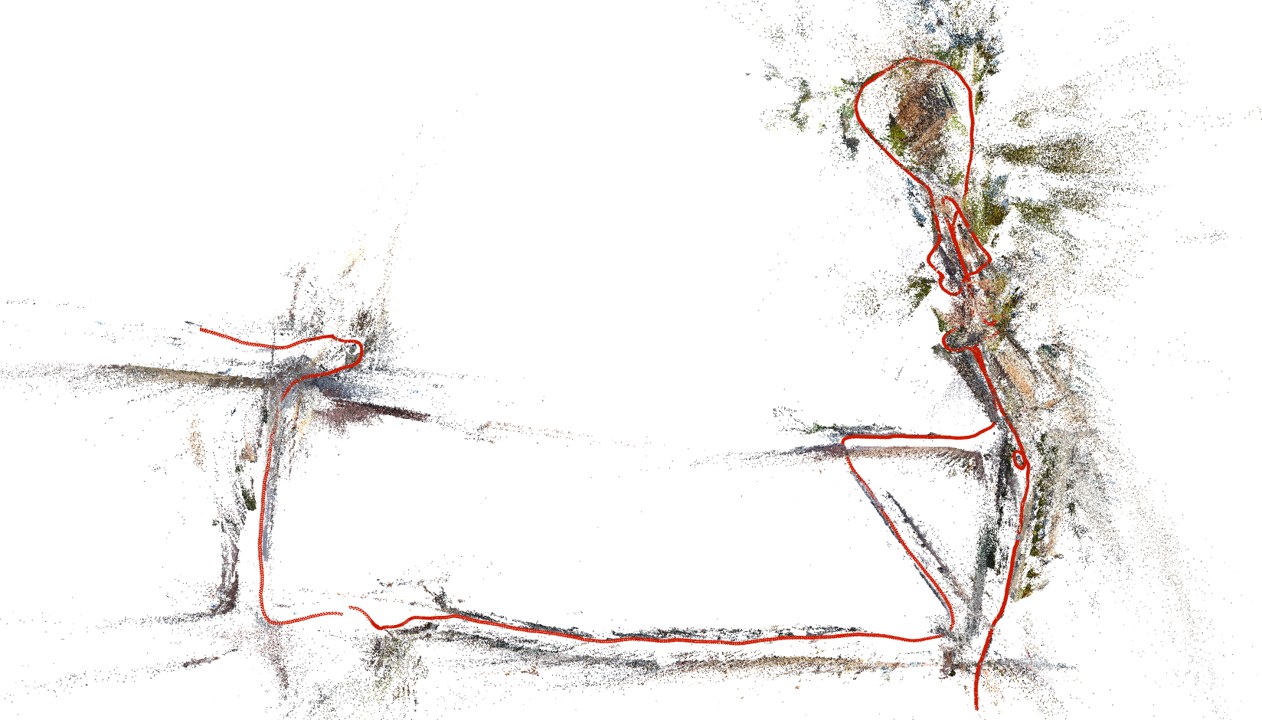} &
 \includegraphics[width=\linewidth]{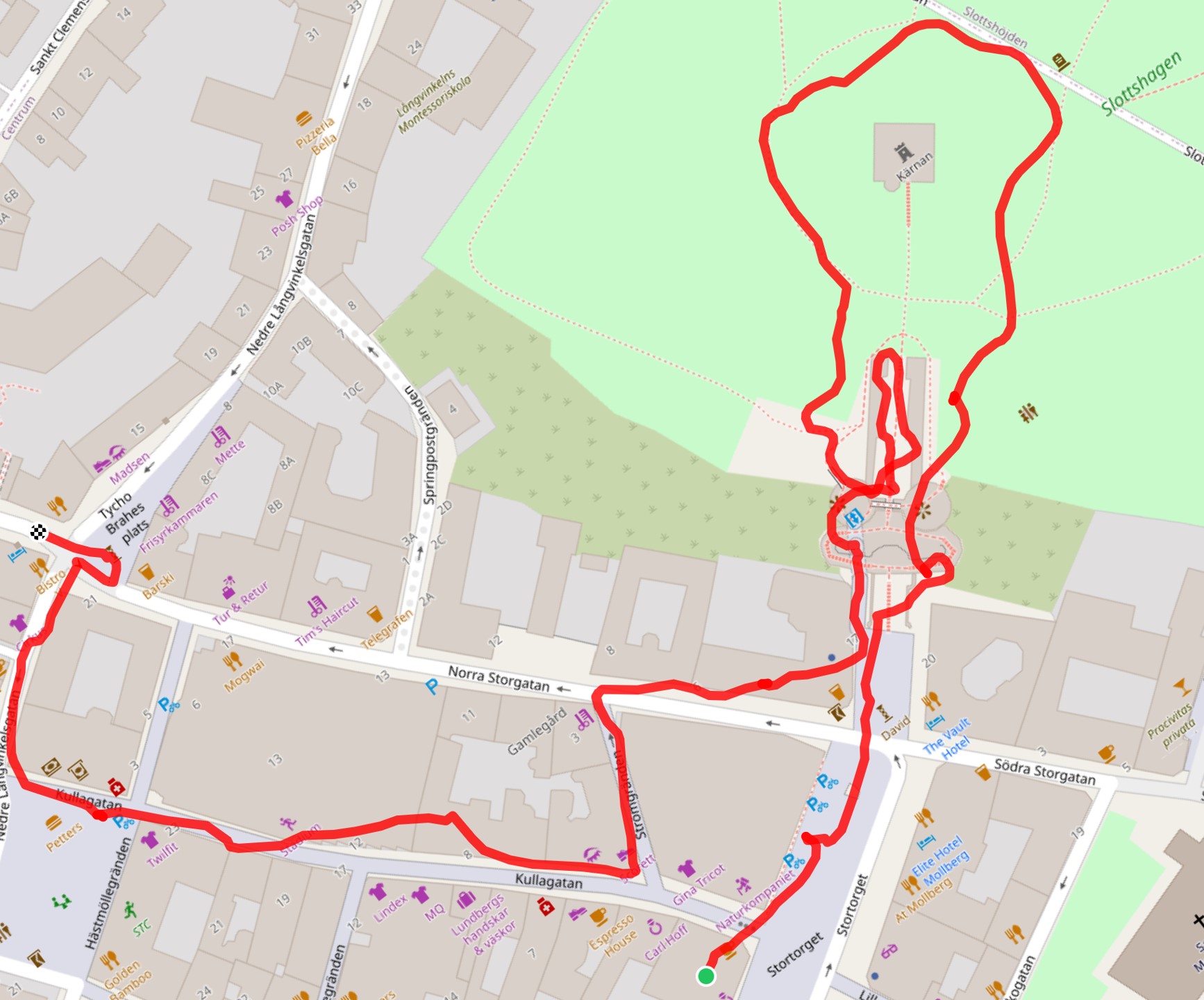}
 \\
 & ~~~~~~~~~~~~~~~~disjoint & ~~~~~~~~~~~~frame missed \& failure & & 
\\

\hdashline

\rotatebox[origin=c]{90}{Helsingborg-2} & \includegraphics[width=\linewidth]{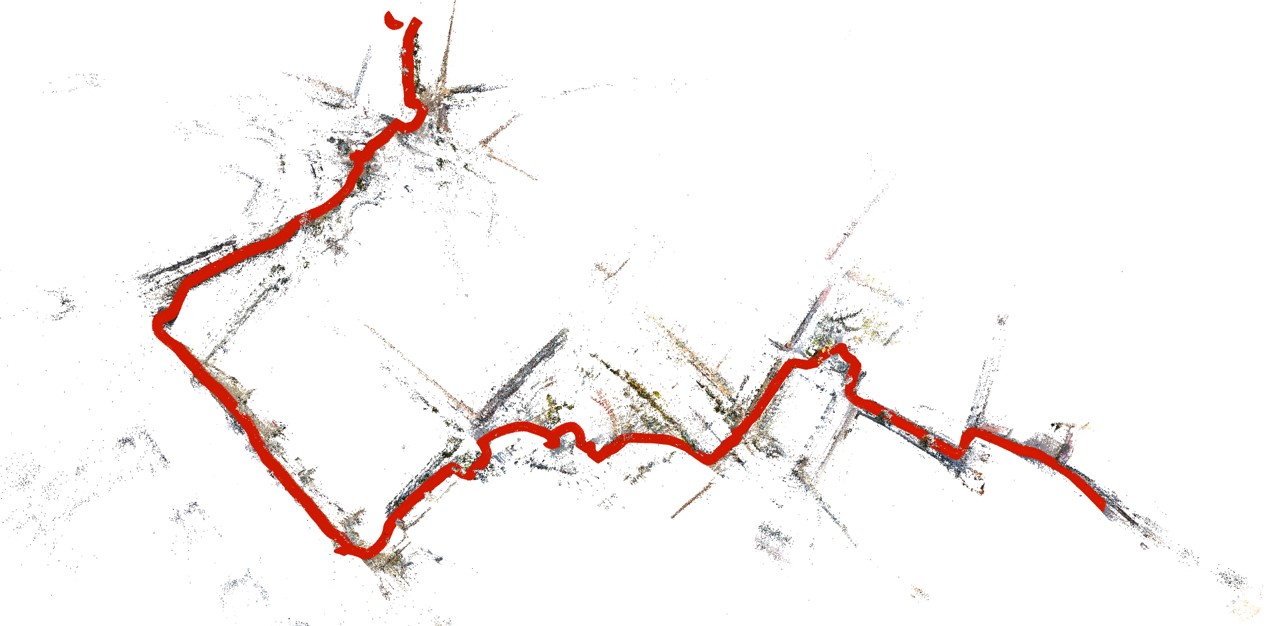}& \includegraphics[width=\linewidth]{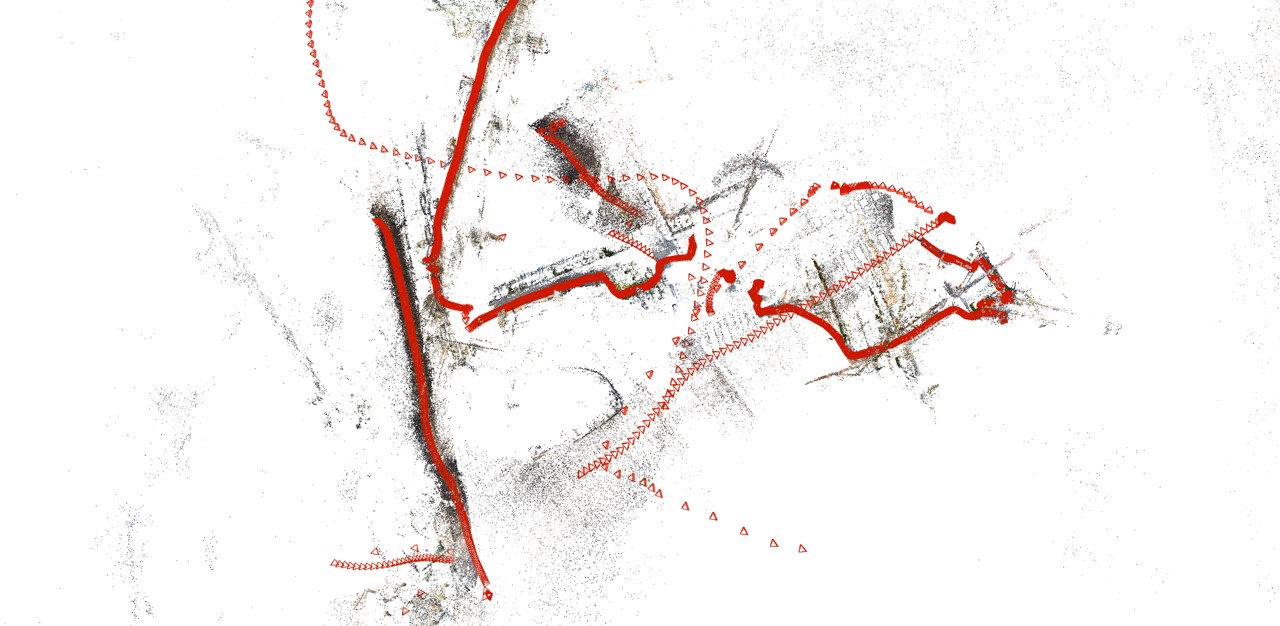}&
\includegraphics[width=\linewidth]{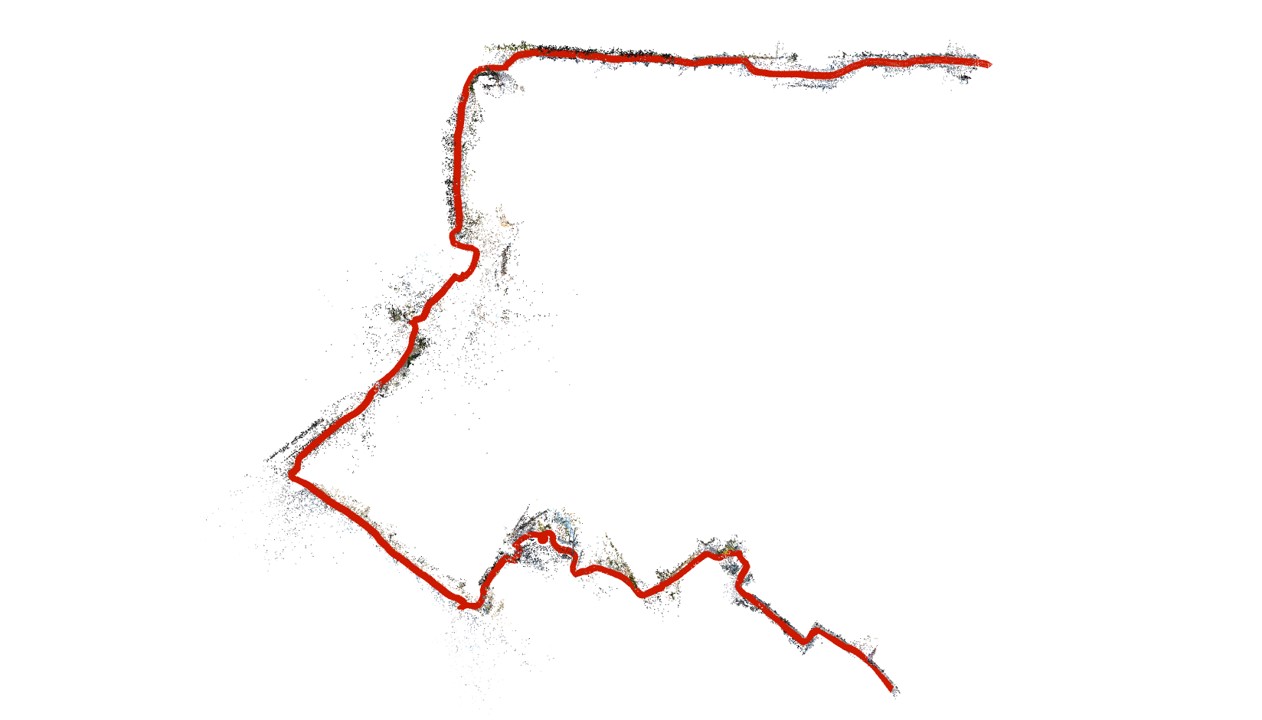} &
\includegraphics[width=\linewidth]{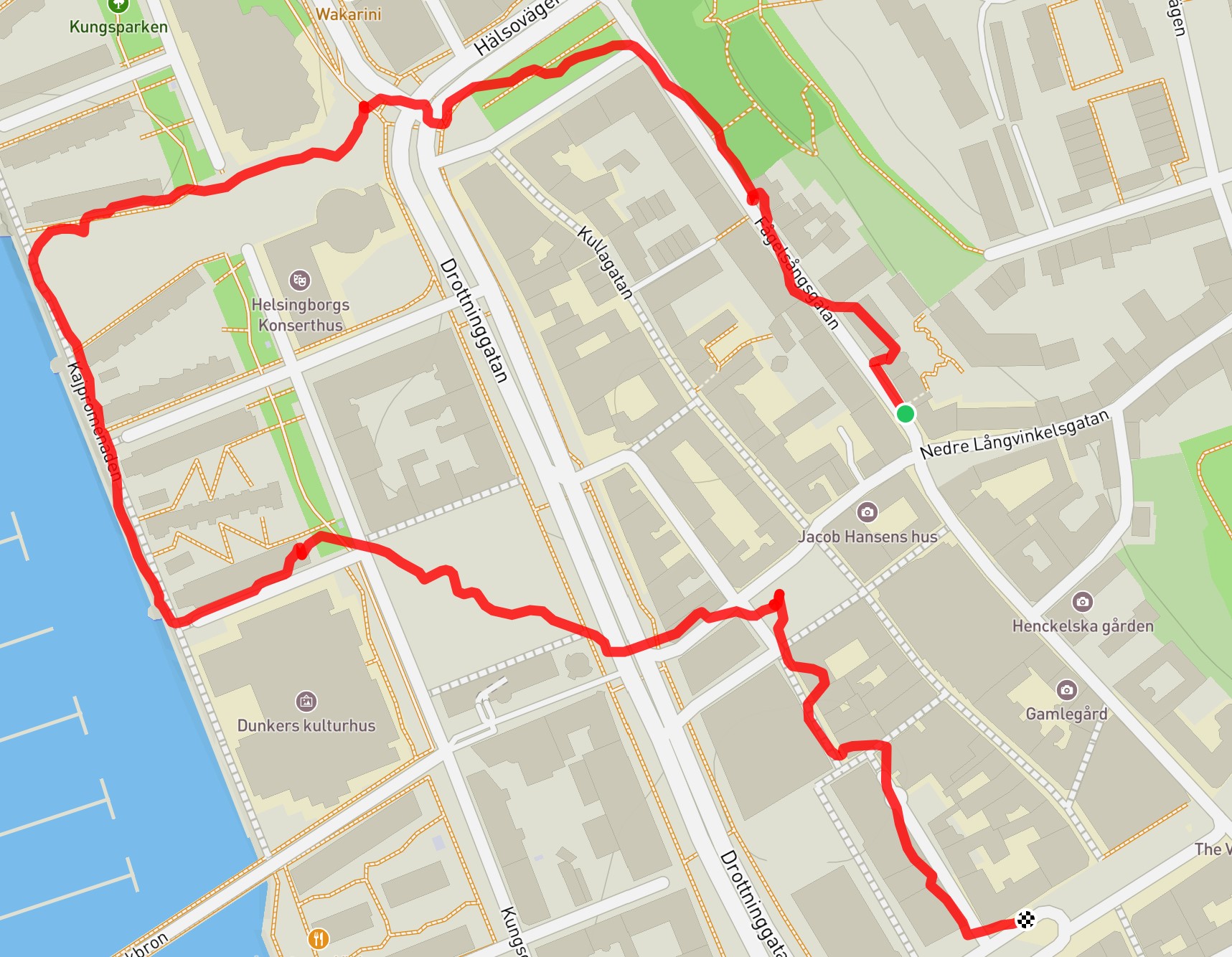}
\\

& ~~~~~~~~~~~~frames missed & ~~~~~~~~~~~~frames missed \& failure & & 
\\

\hdashline

\rotatebox[origin=c]{90}{Lund} & \includegraphics[width=\linewidth]{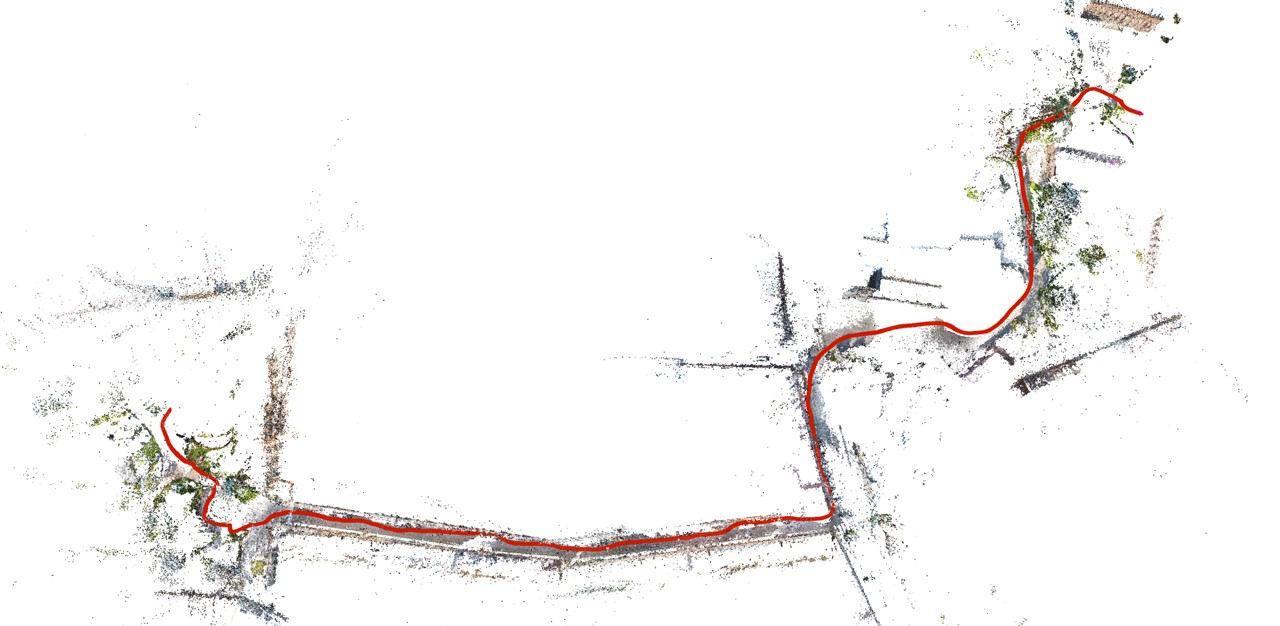}& \includegraphics[width=\linewidth]{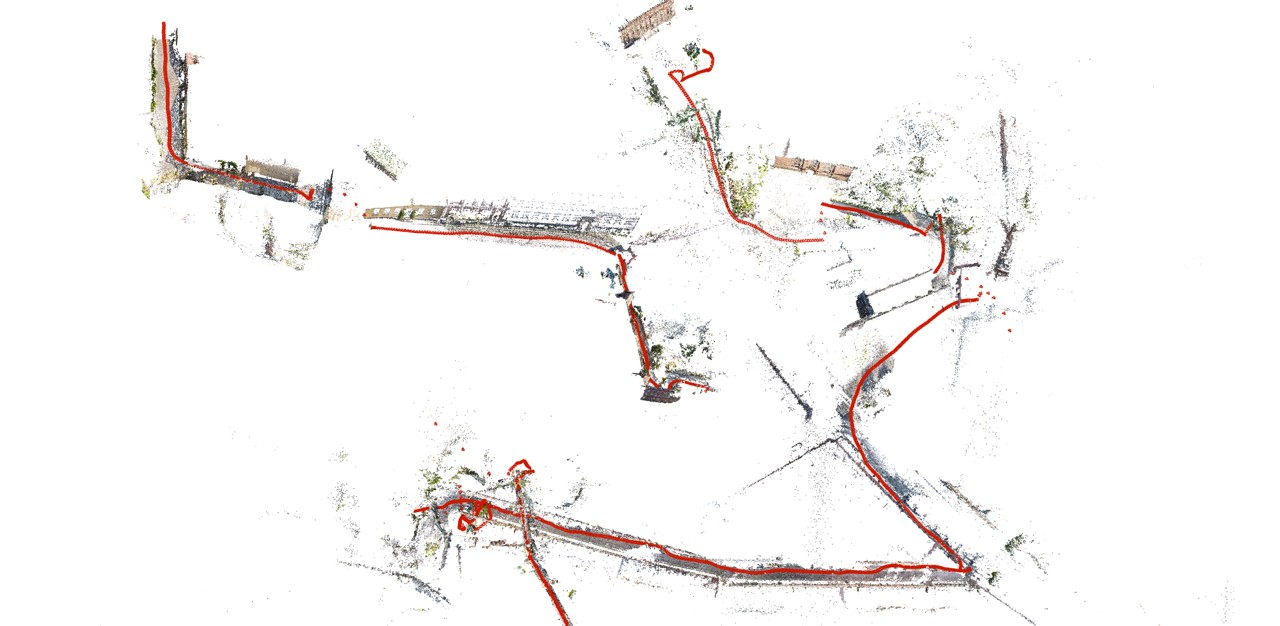}&
\includegraphics[width=\linewidth]{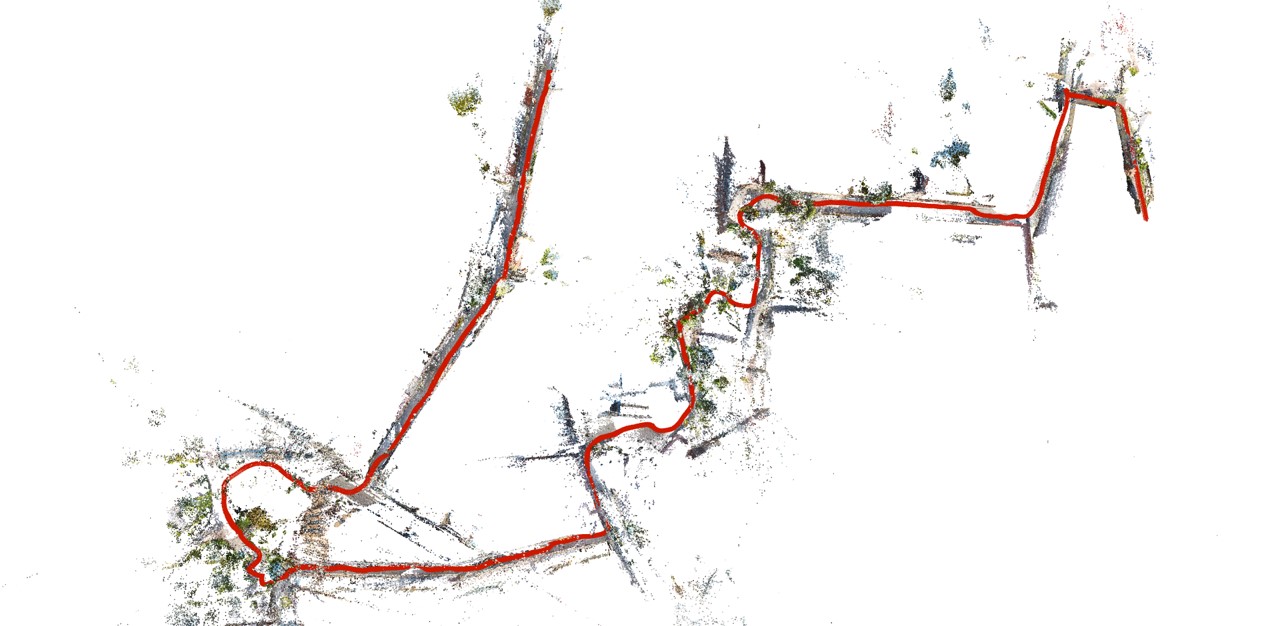} &
\includegraphics[width=\linewidth]{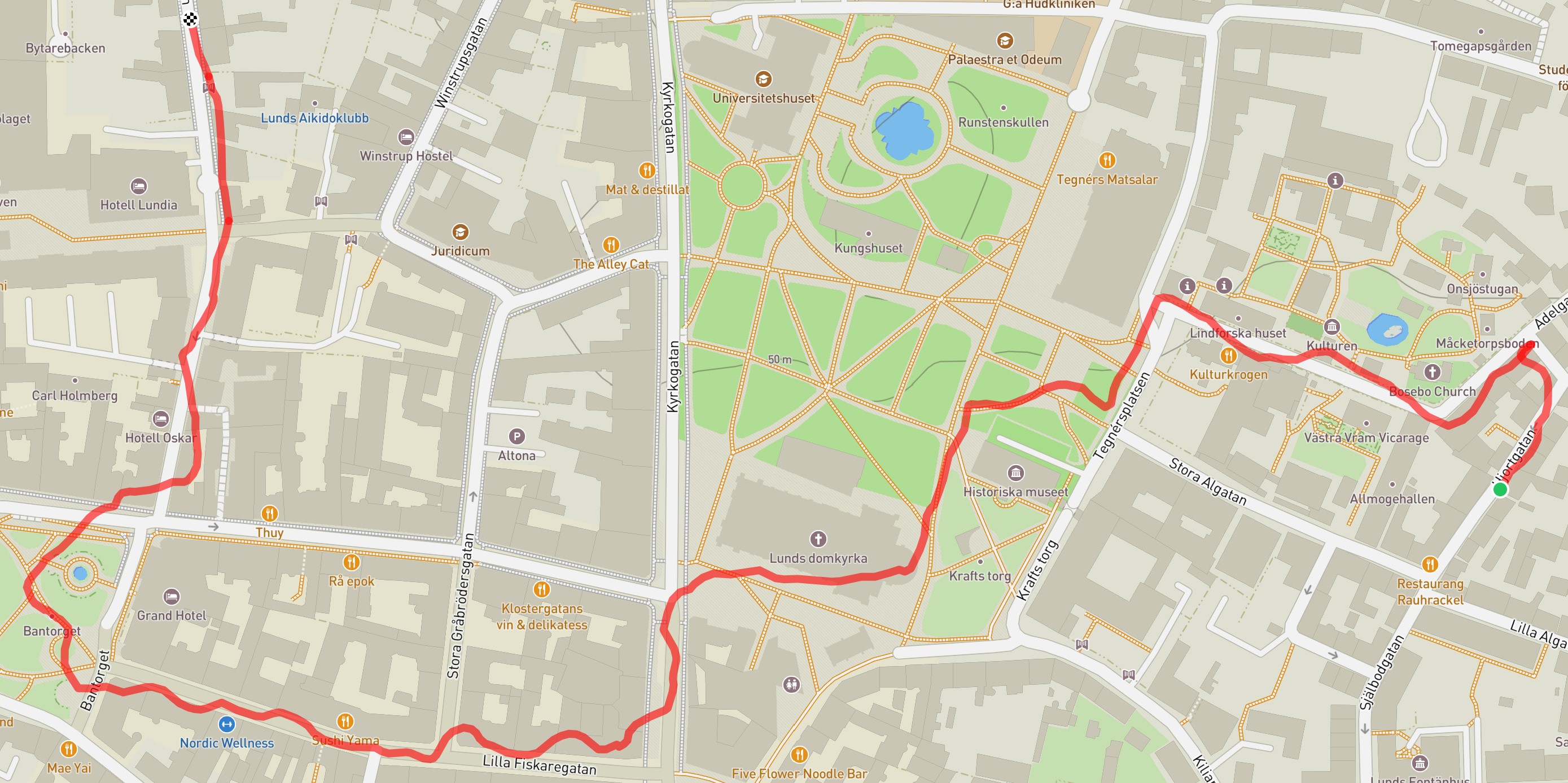}
\\

& ~~~~~~~~~~~~frames missed & ~~~~~~~~~~~~disjoint \& failure & & 
\\

\hdashline

\rotatebox[origin=c]{90}{Uppsala} & \includegraphics[width=\linewidth]{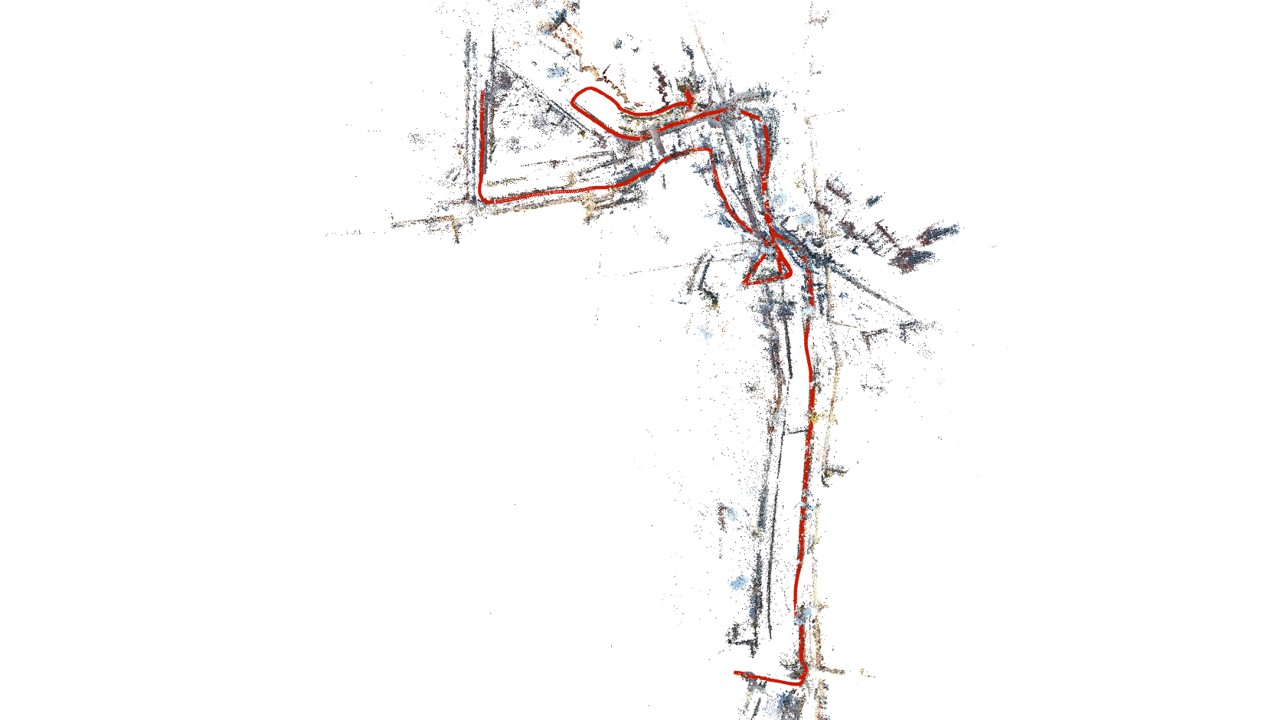}& \includegraphics[width=\linewidth]{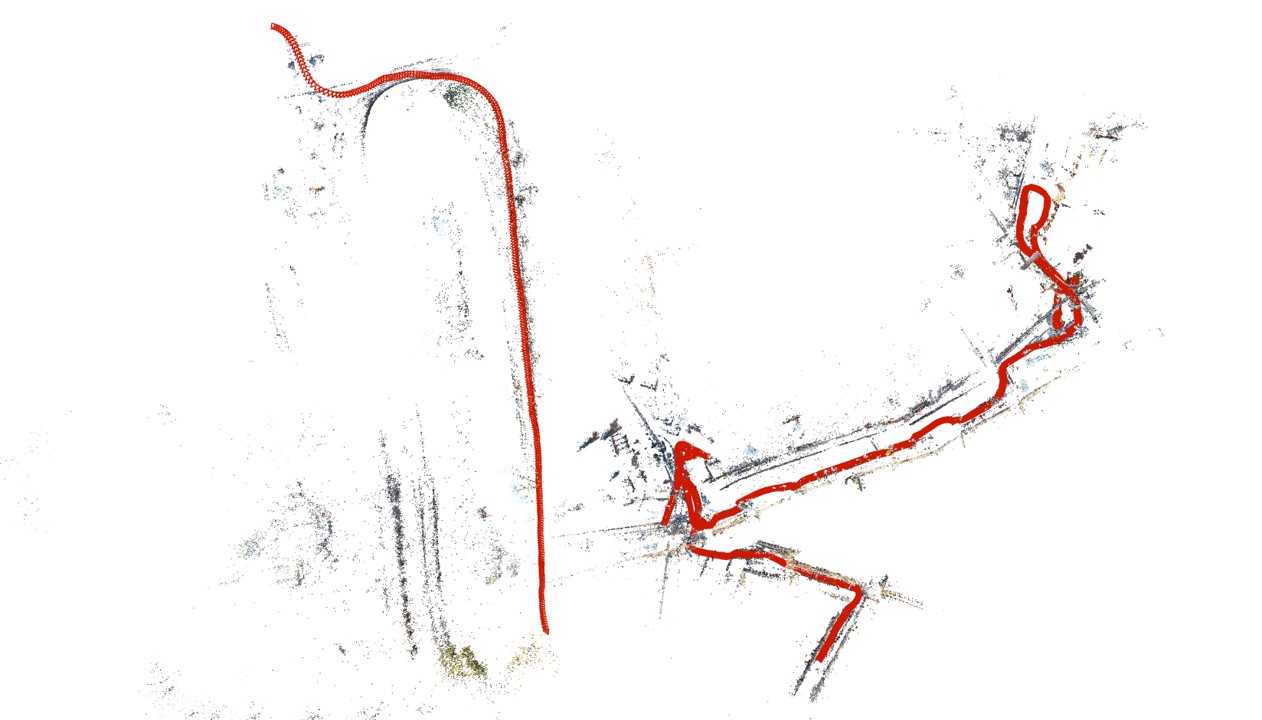}&
\includegraphics[width=\linewidth]{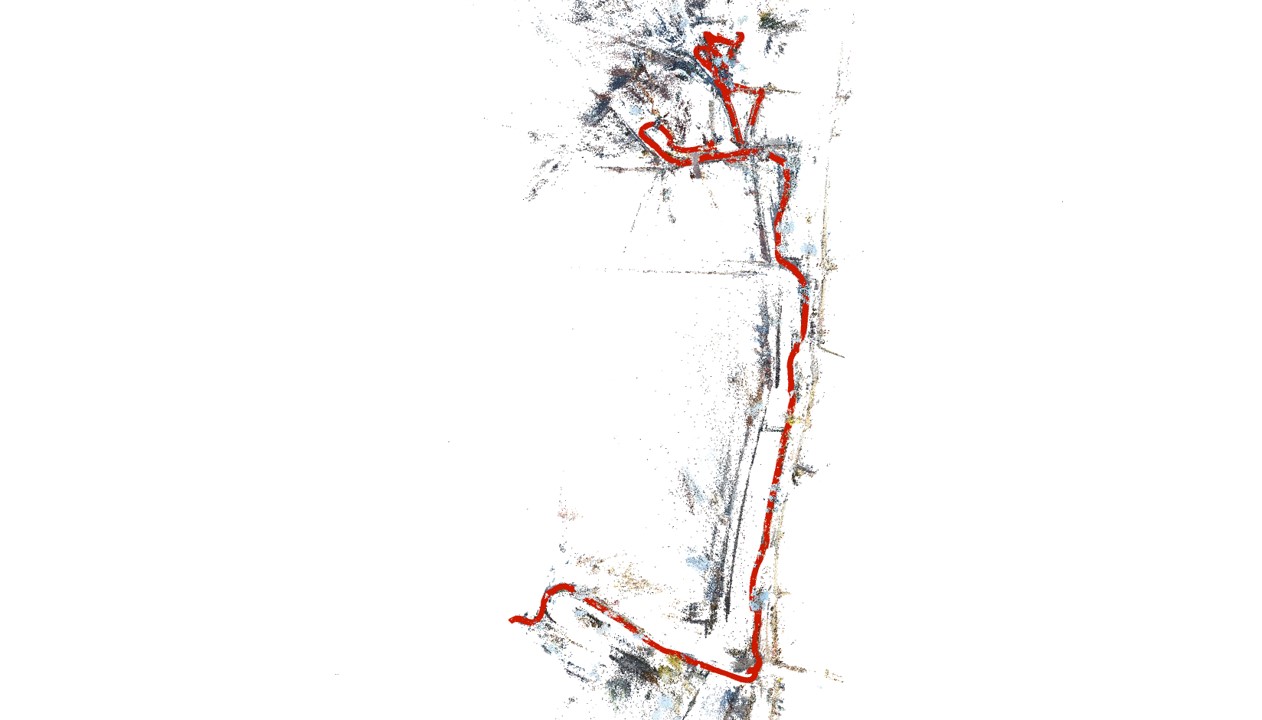} &
\includegraphics[width=\linewidth,clip,trim=0mm 5mm 0mm 5mm]{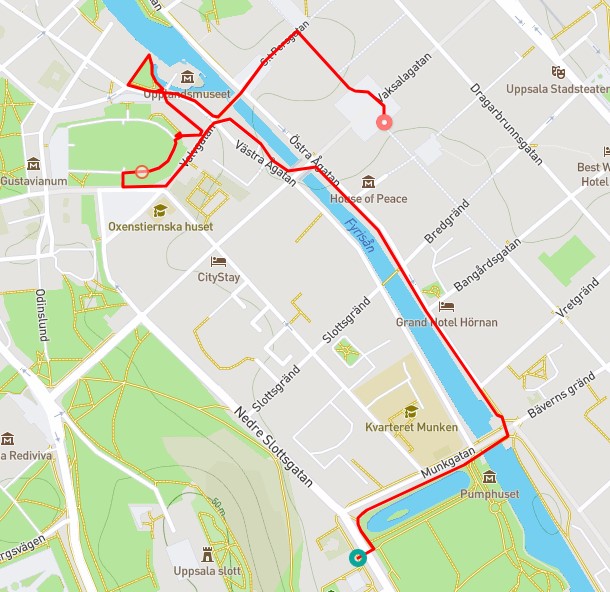}
\\
& ~~~~~~~~~~~~frames missed & ~~~~~~~~~~~~disjoint \& failure & ~~~~~~~~~~~~frames missed & 
\\

\bottomrule

\end{tabular}
}

\caption{\textbf{Reconstruction result visualization}. Our method generally achieves smooth and continuous trajectories without breaks.
COLMAP often produces a model only for part of the path.
GLOMAP struggles to produce consistent results in these large-scale environments.
GPS tracks are provided for Lund and the two Helsingborg sequences, but note that the provided GPS data is rather inaccurate.
For Uppsala, the path has been drawn manually while referencing the video. No reference data is available for Yanshan Park.
}
\label{fig:qualtative_demonstration}
\end{figure*}

%% file: tables/abliation_study.tex
\begin{figure}[t]
    \centering
    \begin{subfigure}{0.23\textwidth}  
        \centering
        \includegraphics[width=\textwidth]{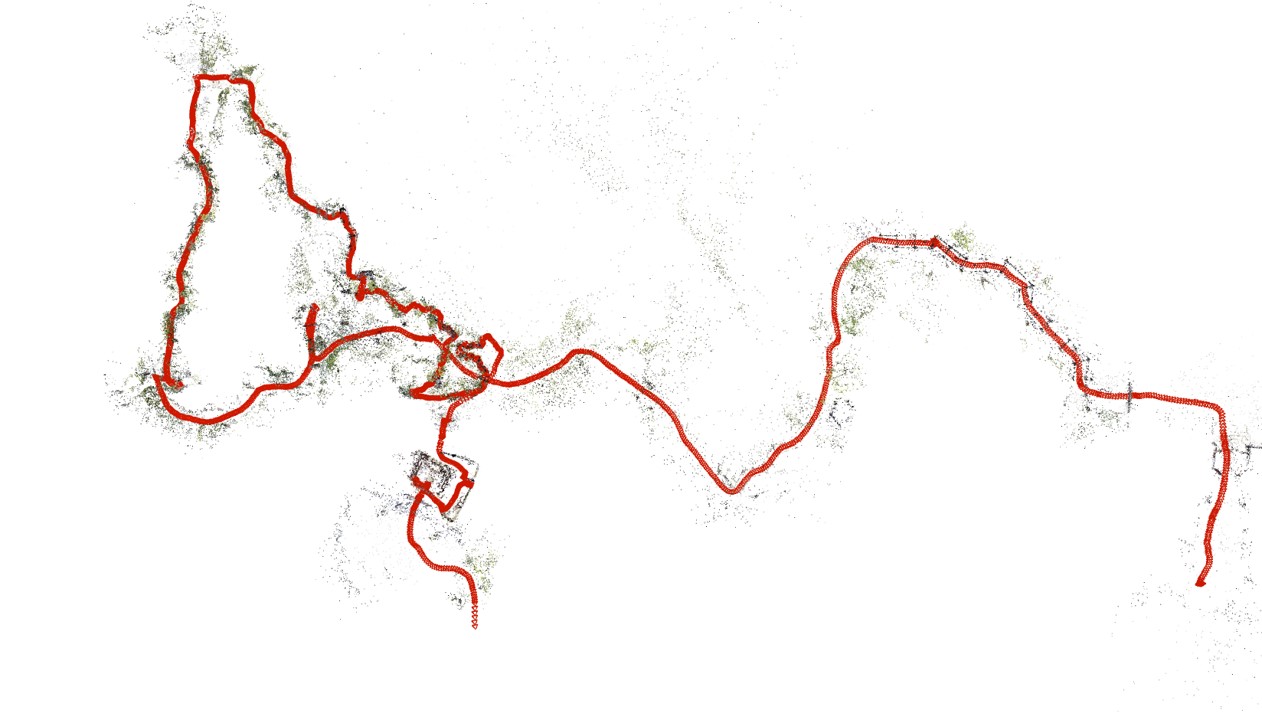}
        \caption{{\color{green!80!black}{\ding{51}}}\rm{Mask}; {\color{red!80!black}{\ding{56}}}Depth;{\color{red!80!black}{\ding{56}}}Loop}
    \end{subfigure}
    \hfill
    \begin{subfigure}{0.23\textwidth}
        \centering
        \includegraphics[width=\textwidth]{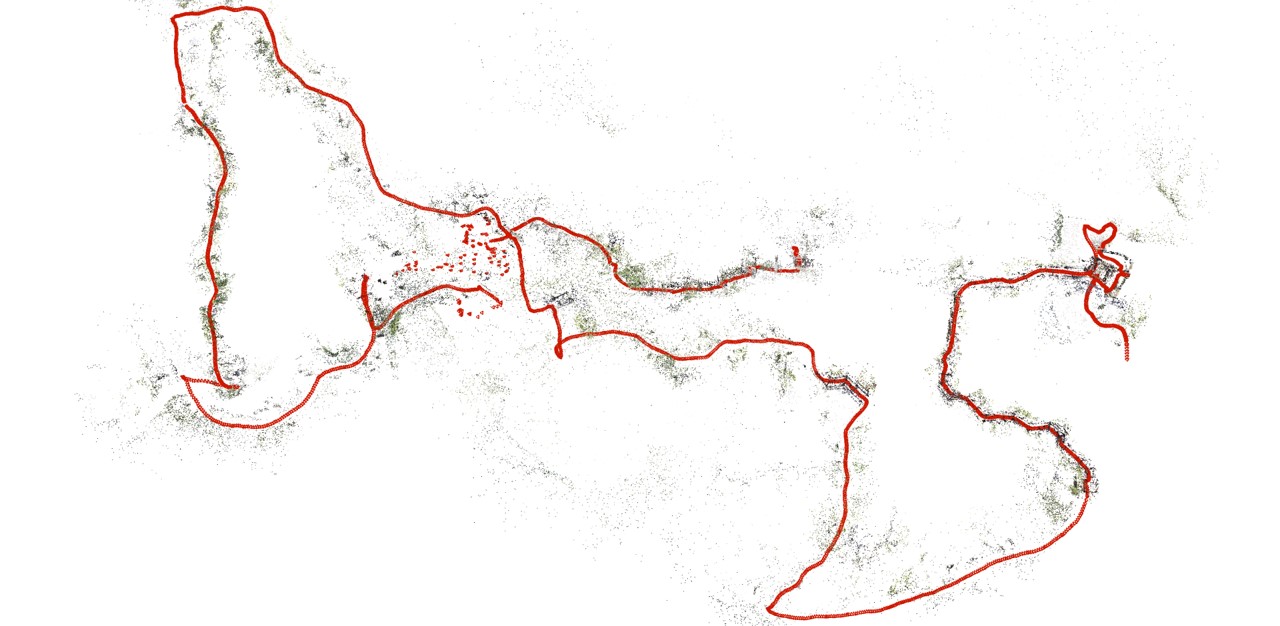}
        \caption{{\color{red!80!black}{\ding{56}}}\rm{Mask}; {\color{green!80!black}{\ding{51}}}Depth;{\color{red!80!black}{\ding{56}}}Loop}
    \end{subfigure}
    
    \vspace{0.2cm}

    \begin{subfigure}{0.23\textwidth}
        \centering
        \includegraphics[width=\textwidth]{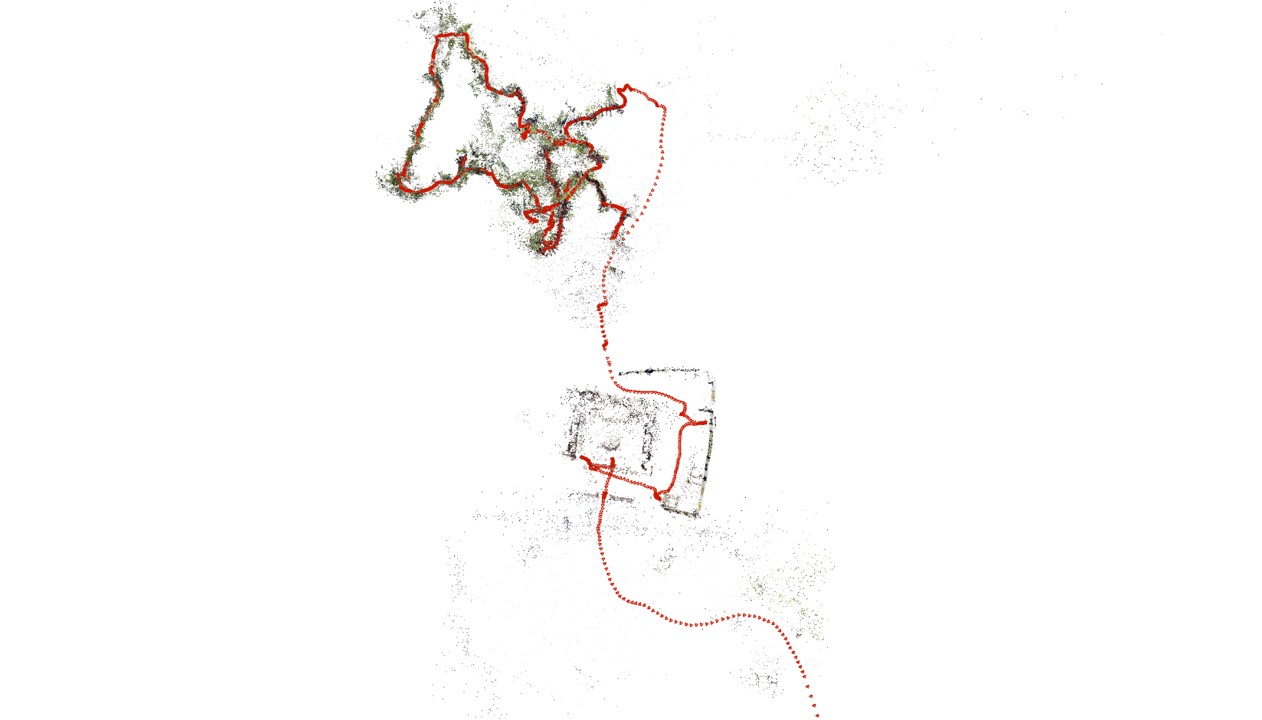}
        \caption{{\color{red!80!black}{\ding{56}}}\rm{Mask}; {\color{red!80!black}{\ding{56}}}Depth;{\color{green!80!black}{\ding{51}}}Loop}
    \end{subfigure}
    \hfill
    \begin{subfigure}{0.23\textwidth}
        \centering
        \includegraphics[width=\textwidth]{figures/yanshanyuan/ours_park2.jpg}
        \caption{{\color{green!80!black}{\ding{51}}}\rm{Mask}; {\color{green!80!black}{\ding{51}}}Depth;{\color{green!80!black}{\ding{51}}}Loop}
    \end{subfigure}

    \caption{\textbf{Ablation study}: on the ``Yanshan Park" sequence, we show the effectiveness of the proposed modules.
    To handle in-the-wild videos, prior depth and pruning dynamics in the view greatly help to improve the robustness.}
    \label{fig:abliation_study}
\end{figure}